\documentclass[authoryear,preprint,review,12pt]{elsarticle}

\usepackage{amssymb}
\usepackage{amsmath}
\usepackage{url}
\usepackage{booktabs}
\usepackage{multirow}
\usepackage{multicol}

\usepackage{soul}
\usepackage{xcolor}

\usepackage{lineno}

\journal{}

\begin{document}

\begin{frontmatter}



% \Title{Improved implicit diffusion model with knowledge distillation to estimate the spatial distribution density of carbon stock in remote sensing imagery}
% \Title{Estimating forest carbon stocks from medium resolution remote sensing imagery with image-to-image translation method}
% \Title{IIDM: Improved Implicit Diffusion Model with Knowledge Distillation for Carbon Stock Density Estimation in Remote Sensing Imagery}
\title{IIDM: Improved Implicit Diffusion Model with Knowledge Distillation to Estimate the Spatial Distribution Density of Carbon Stock in Remote Sensing Imagery}


\author[label1]{Zhenyu Yu} %% Author name
\ead{yuzhenyuyxl@foxmail.com}
% \author[label1]{Chee Seng Chan\corref{cor1}}
% \ead{cs.chan@um.edu.my}
% \author[label2]{Pei Wang\corref{cor1}}
% \ead{peiwang@mail.ynu.edu.cn}
\author[label2]{Jinnian Wang}
\ead{jnwang@gzhu.edu.cn}
\author[label1]{Mohd Yamani Idna Idris\corref{cor1}}
\ead{yamani@um.edu.my}
\cortext[cor1]{Corresponding author.}
% \newcommand{\orcidauthorA}{0000-0002-9985-0165} 

% Faculty of Computer Science and Information Technology, Universiti Malaya, Kuala Lumpur, 50603, Malaysia; yuzhenyuyxl@foxmail.com\\
% $^{2}$ \quad Faculty of Computer Science and Information Technology, Universiti Malaya, Kuala Lumpur, 50603, Malaysia; cs.chan@um.edu.my\\

\affiliation[label1]{organization={Faculty of Computer Science and Information Technology},%Department and Organization
            addressline={Universiti Malaya}, 
            city={Kuala Lumpur},
            postcode={50603}, 
            state={Kuala Lumpur},
            country={Malaysia}}
% \affiliation[label2]{organization={Faculty of Information Engineering and Automation},%Department and Organization
%             addressline={Kunming University of Science and Technology}, 
%             city={Kunming},
%             postcode={650504}, 
%             state={Yunnan Province},
%             country={China}}

\affiliation[label2]{organization={Faculty of Geography and Remote Sensing},%Department and Organization
            addressline={Guangzhou University}, 
            city={Guangzhou},
            postcode={510006}, 
            state={Guangdong Province},
            country={China}}

%% Abstract
\begin{abstract}
Forests play a critical role as terrestrial carbon sinks, effectively reducing atmospheric CO$_2$ concentrations and mitigating climate change. Remote sensing facilitates large-scale, high-accuracy carbon stock estimation, with optical imagery enabling long-term monitoring. This study focuses on Huize County, Qujing City, Yunnan Province, China, utilizing GF-1 WFV satellite imagery. We propose an Improved Implicit Diffusion Model (IIDM), integrating Knowledge Distillation (KD) for feature extraction and spatial distribution modeling. 
The key contributions are as follows: (1) A knowledge distillation-enhanced VGG (KD-VGG) module for efficient feature extraction, reducing inference time while preserving crucial spectral and spatial information. (2) A KD-UNet architecture with cross-attention and MLP modules for improved feature fusion, effectively capturing global-local relationships and enhancing estimation accuracy. (3) A novel IIDM framework incorporating generative diffusion modeling and implicit neural representations, enabling high-resolution spatial carbon stock estimation. IIDM achieves state-of-the-art accuracy, with an RMSE of 12.17\%, outperforming traditional regression models by 41.69\% to 42.33\%. The model effectively extracts deep spatial and spectral features, surpassing existing methods and reinforcing the feasibility of AI-generated content in quantitative remote sensing. The high-resolution (16-meter) carbon stock distribution maps provide a solid foundation for refining forest carbon sink regulations and improving regional carbon stock management strategies.
\end{abstract}


\begin{highlights}
\item \textbf{KD-VGG and KD-UNet models} significantly improve feature extraction accuracy and reduce inference time through knowledge distillation.

\item \textbf{Cross-attention + MLPs module} effectively fuses global and local features, leading to superior carbon stock estimation accuracy.

\item \textbf{Improved Implicit Diffusion Model (IIDM)} achieves 41.69\%-42.33\% higher accuracy than traditional regression models, with an RMSE of 12.17\%.

\end{highlights}

%% Keywords

\end{frontmatter}

% \linenumbers


\section{Introduction}

Given the increasing threat of global climate change, human activities pose unprecedented challenges to Earth's ecosystems \cite{lenoir2020species}. As a result, forest resource management and utilization have become primary concerns for researchers and policymakers worldwide \cite{pecchi2019species}. {Forest ecosystems play a pivotal role in regulating the global carbon cycle, offering critical ecological services and contributing significantly to climate change mitigation} \cite{brando2019droughts}. {Accurately estimating the spatial distribution and temporal variation of regional carbon stocks is essential for understanding forest carbon dynamics and formulating effective regulatory strategies} \cite{nolan2021constraints, hua2022biodiversity}.

Recent advancements in remote sensing technology have facilitated large-scale forest carbon stock monitoring by integrating ground-based sample data with satellite observations \cite{hurtt2019beyond, sun2020review, lee2021first, santoro2022dynamics}. While ground surveys provide high-precision data, they are labor-intensive, time-consuming, and impractical for large-scale monitoring \cite{xiao2019remote, gao2023spatially}. In contrast, {remote sensing-based inversion models provide efficient and scalable solutions for estimating carbon stocks across large regions} \cite{long2020high, gray2020remote}. Conventional spectral information-based methods infer vegetation conditions from optical imagery to estimate carbon stocks, but they struggle to model complex nonlinear relationships due to variations in image quality, spectral distortions, and spatial resolution limitations \cite{chen2019remote}.

Structural information-based methods measure forest biomass and carbon stock more directly, but their effectiveness is constrained by the spatial resolution and coverage of remote sensing images \cite{xiao2019remote}. Physical model-based approaches utilize ecosystem carbon-cycle simulations, offering theoretical insights but requiring {extensive field-measured ecological parameters, making them challenging to generalize for large-scale applications} \cite{he2021soil}. Machine learning (ML)-based models have emerged as a powerful alternative, revealing {hidden patterns in remote sensing data and improving estimation accuracy without reliance on predefined physical assumptions} \cite{odebiri2021basic, wang2022using}. However, traditional ML models such as regression and decision tree-based algorithms often fail to fully capture the spatial dependencies and hierarchical feature relationships within remote sensing images.

{To address these limitations, we introduce the Improved Implicit Diffusion Model (IIDM), a novel framework that integrates diffusion models with implicit neural representations for enhanced carbon stock estimation. IIDM leverages a generative approach to refine spatial distribution density estimation, surpassing conventional ML techniques in extracting deep spatial features. By incorporating knowledge distillation, our method optimizes feature extraction while reducing model complexity, thereby improving both efficiency and accuracy.}

{This study applies IIDM to Huize County, using GF-1 WFV satellite imagery as the primary data source to achieve high-precision regional carbon stock estimations. Our approach provides a theoretical foundation for improved forest carbon sink regulations and data-driven decision-making in climate change mitigation. The main contributions of our work are as follows:}

\begin{itemize} \item {We integrate knowledge distillation into the VGG module within the diffusion model, enhancing initial feature extraction. Additionally, UNet undergoes distillation, leading to a significant reduction in parameters while preserving feature representation, thereby improving inference efficiency.} \item {We introduce a cross-attention mechanism combined with MLPs to enhance feature fusion, effectively capturing complex global-local dependencies and improving the spatial distribution accuracy of carbon stock density estimation.} \item {We propose the IIDM framework, which optimizes the inference process while achieving state-of-the-art estimation accuracy. This demonstrates the feasibility of leveraging artificial intelligence-generated content (AIGC) for high-precision remote sensing applications.} \end{itemize}

\section{Related Work}

\subsection{Key Issues in Carbon Stock Estimation}

\textbf{Definition and Importance.} {Carbon stock refers to the total amount of carbon stored in a forest ecosystem, including vegetation, soil, and biomass. It plays a crucial role in global climate regulation by acting as a carbon sink that absorbs atmospheric CO$_2$. However, estimating carbon stock with high spatial resolution remains challenging due to environmental heterogeneity and the limitations of existing remote sensing techniques.} Understanding the \textbf{spatial distribution density} of carbon stock is essential for precise forest resource management, carbon trading, and climate change mitigation policies.

\textbf{Data Selection.} Ground monitoring provides precise data but is impractical for large-scale observations due to its labor-intensive and time-consuming nature \cite{xiao2019remote, gao2023spatially}. Remote sensing inversion methods mitigate these issues, improving both efficiency and accuracy \cite{long2020high, gray2020remote}. Among the most promising remote sensing resources for additional datasets are medium-resolution (10$\sim$30 m) optical data. The long operational lifetimes of these satellites make them ideal for continuous monitoring of forest dynamics \cite{puliti2021above}. However, {optical multispectral data, though widely used, suffer from spectral band limitations that hinder their ability to capture fine-scale variations in carbon stock. This challenge necessitates the adoption of deep learning models capable of extracting hierarchical feature representations to improve estimation accuracy.}

\textbf{Method Selection.} Traditional carbon stock estimation methods integrate field observations with remote sensing imagery. Zhang et al. \cite{zhang2019mapping} combined MODIS, GLAS, climate, and terrain data with random forests to produce a 1 km resolution biomass map for China, achieving an interpretability of 75\% and an RMSE of 45.5 Mg/ha. Puliti et al. \cite{puliti2021above} estimated forest biomass change over 1.4 million hectares in Norway using Sentinel-2 and Landsat data. Chopping et al. \cite{chopping2022forest} used MISR to estimate biomass in the southwestern U.S. from 2000 to 2015, achieving an RMSE of 37.0 Mg/ha. {While these methods offer valuable insights, their reliance on traditional machine learning models limits their ability to capture intricate spatial dependencies and nonlinear feature relationships critical for precise carbon stock estimation.} 

{To overcome these challenges, we introduce the Improved Implicit Diffusion Model (IIDM), a generative deep learning approach that enhances carbon stock estimation by leveraging feature learning techniques, knowledge distillation, and multi-scale spatial feature extraction.}

\subsection{Challenges in Quantitative Remote Sensing}

One of the most significant challenges in \textbf{quantitative remote sensing} is effectively managing complex non-linear relationships while improving estimation accuracy \cite{qrs_han2023survey, qrs_yasir2023coupling}. Although remote sensing inversion methods allow efficient monitoring of forest carbon stocks, obstacles such as low image quality, spectral noise, and environmental heterogeneity still hinder their performance. {Current methods struggle to generalize across diverse ecological conditions, necessitating advanced deep learning models capable of dynamically adapting to different environmental settings.} 

This study introduces {IIDM, which integrates implicit neural representations into the diffusion framework to refine carbon stock estimation, effectively addressing issues related to image degradation, data sparsity, and inconsistent spectral responses.} 

Deep learning surpasses traditional machine learning methods in handling high-dimensional data and extracting hierarchical representations, making it particularly effective for remote sensing applications \cite{qrs_wang2023review, qrs_han2023survey}. {Unlike traditional models that rely on manually selected features, IIDM employs an end-to-end learning paradigm that automatically refines spatial feature representations, improving accuracy and robustness in carbon stock estimation.}

\subsection{The Intersection of Quantitative Remote Sensing and Computer Vision}

Recent advancements in deep learning have significantly enhanced feature extraction in remote sensing applications \cite{xiao2019remote, matinfar2021evaluation, pham2023advances}. {In particular, the fusion of deep learning and artificial intelligence-generated content (AIGC) has introduced a new paradigm for high-precision carbon stock estimation. This approach reconstructs spatial distribution density with enhanced accuracy, leveraging generative models to synthesize missing information and refine predictions. Our proposed IIDM follows this paradigm by employing a diffusion-based generative framework to enhance spatial feature representation and estimation accuracy.}

\textbf{Machine Learning.}  
Many current studies have utilized traditional machine learning methods such as Ordinary Least Squares (OLS), Random Forest (RF), and Support Vector Regression (SVR) for carbon stock estimation \cite{li2024review, qrs_han2023survey, qrs_wang2023review}. While effective for linear and moderately nonlinear relationships, these models often fail to generalize in heterogeneous landscapes. {Unlike these traditional approaches, IIDM leverages deep feature learning to model complex nonlinear dependencies between remote sensing data and carbon stock distribution. Furthermore, IIDM benefits from the knowledge distillation framework, which significantly improves efficiency while maintaining feature representational power, addressing the high computational demands often associated with deep learning-based estimation models.}

\textbf{Generative Models.}  
The Generative Adversarial Network (GAN), proposed by Goodfellow et al. \cite{goodfellow2014generative}, employs adversarial training to synthesize realistic samples. However, GAN training is inherently unstable, prone to mode collapse, and requires extensive hyperparameter tuning. Variational Autoencoders (VAE), introduced by Kingma and Welling \cite{kingma2013auto}, improve sample diversity but often produce blurry reconstructions. {IIDM overcomes these limitations by utilizing an implicit diffusion framework, which refines feature representations progressively, ensuring stable and accurate carbon stock estimations. Unlike GANs, which require adversarial optimization, and VAEs, which struggle with sample sharpness, diffusion models provide a principled approach to generating high-quality feature distributions.}

\textbf{Diffusion Models.}  
The diffusion model, first introduced by Sohl-Dickstein et al. \cite{sohl2015deep}, progressively removes Gaussian noise from training images, functioning similarly to denoising autoencoders. Latent Diffusion Models (LDMs) refine this approach by mapping images into compact latent spaces. Unlike GANs, diffusion models offer stable training, improved sample diversity, and resilience to mode collapse. 
{Our proposed IIDM builds upon the strengths of diffusion models while addressing their computational inefficiencies through knowledge distillation and multi-scale feature extraction. Traditional diffusion models, such as }DDPM \cite{ho2020denoising}, DDIM \cite{song2020denoising}, and Stable Diffusion \cite{rombach2022high}, {focus on generating high-quality image samples, whereas IIDM is explicitly designed for structured regression tasks, such as carbon stock estimation, by integrating deep hierarchical feature representations and an optimized implicit neural framework.}

{In summary, IIDM bridges the gap between conventional machine learning, generative models, and diffusion models by providing a novel knowledge distillation-enhanced diffusion framework tailored for high-precision spatial carbon stock estimation. This represents a significant step forward in AI-driven remote sensing and quantitative environmental modeling.}

\section{Materials}

\subsection{Study area}
\begin{figure}[!ht]
\centering
\includegraphics[width=1.0\textwidth]{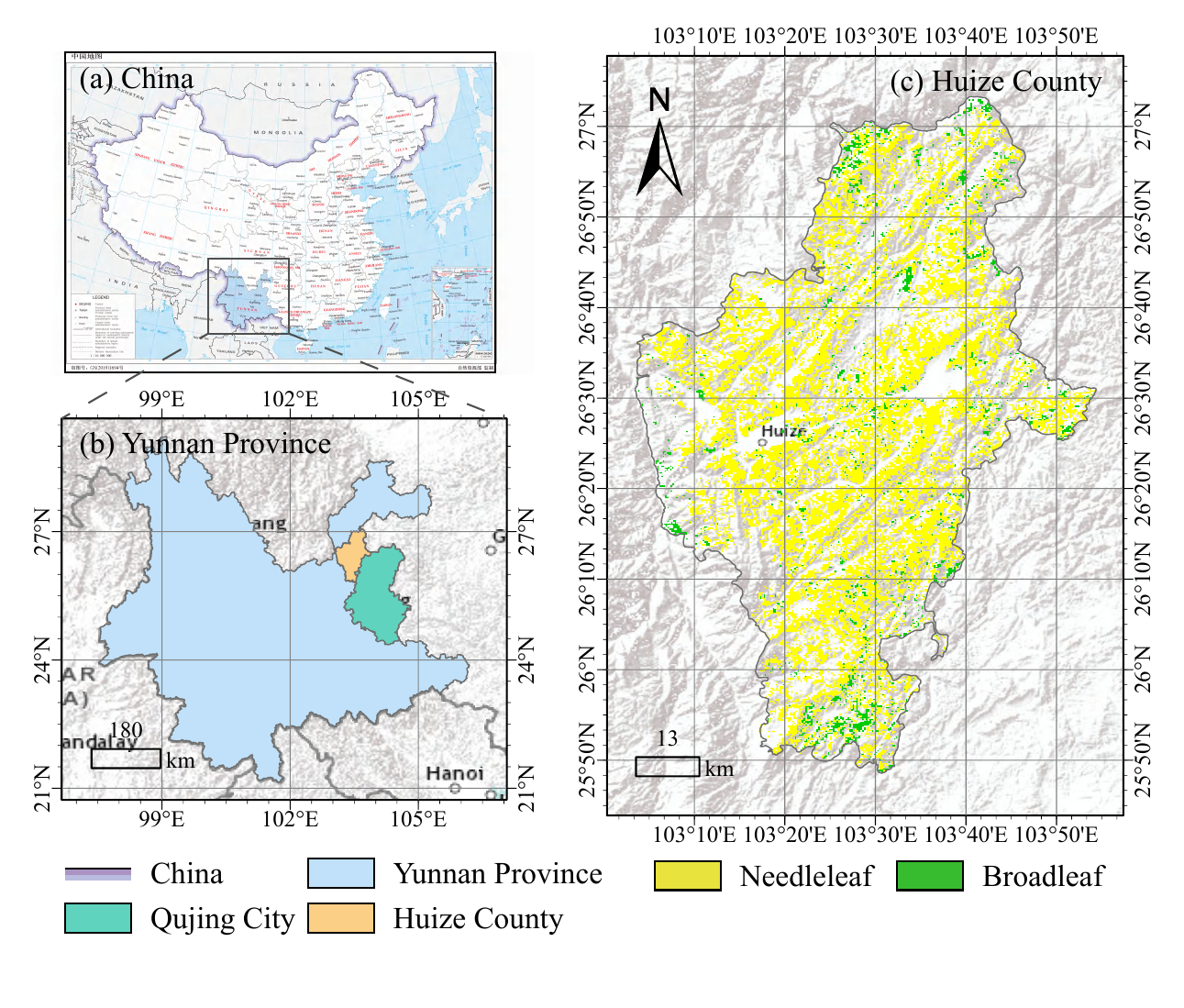}
\caption{Study area. Notes: (a) China, (b) Yunnan Province, and (c) Huize County. The basemap is provided by ArcGIS Pro.}
\label{fig1}
\end{figure}

Huize County, located in Qujing City, Yunnan Province, China, covers an area of 5,889 km$^2$ (Figure \ref{fig1}). The county features a ladder-shaped terrain, with elevations ranging from the highest peak at 4,017 m to the lowest point at 695 m. {This complex topography significantly influences forest carbon distribution and presents challenges for remote sensing-based carbon stock estimation.} The region experiences a temperate plateau monsoon climate, transitioning from a subtropical climate in the south to a cold temperate climate in the north, {which further impacts vegetation growth patterns and carbon sequestration capacity.} According to Forest Management Inventory data, Huize County has approximately 3,080.53 km$^2$ of forest land, with arbor forest land covering approximately 2,538.20 km$^2$ (82.39\%). {The rich forest resources and diverse ecological conditions make Huize County an ideal site for carbon stock estimation research, despite the challenges posed by its complex geographical landscape.}

\subsection{Data sources}
{This study integrates multiple data sources, including field survey data, satellite imagery, and topographic datasets, to enhance carbon stock estimation accuracy.}

\textbf{Survey data.} The study utilized 2020 data from the Forest Management Inventory obtained from the Huize County Forestry and Grass Bureau. This dataset comprises over 70 attributes, such as forest area and type, {enabling a comprehensive analysis of forest distribution and carbon storage dynamics.} These data provide a scientific basis for forestry management, protection, and sustainable utilization.

\textbf{Satellite imagery.} The study primarily used GF-1 WFV image data. Launched by China in 2013, the GF-1 satellite provides medium-resolution remote sensing with a spatial resolution of 16 meters. {It features four spectral bands (red, green, blue, and near-infrared), which are essential for vegetation monitoring and carbon stock estimation.} To ensure time consistency, images from August 27, 2020, were used for model training, covering the entire study area. Data details are provided in Table A1.

\textbf{Elevation data.} The study utilized ALOS PALSAR DEM data with a spatial resolution of 12.5 meters, achieving full coverage of the study area with four-view images (Table A1). ALOS PALSAR, a satellite equipped with a Synthetic Aperture Radar (SAR) sensor, {provides detailed, all-weather topographic measurements, which are crucial for understanding terrain-induced variations in carbon stock distribution.}

\textbf{Canopy height.} The study utilized a 10-meter vegetation canopy height dataset from the ETH Global Sentinel-2 \cite{lang2022high}. Lang et al. \cite{lang2022high} fused GEDI and Sentinel-2 data to develop a probabilistic deep learning model for retrieving canopy height from Sentinel-2 imagery globally. The global canopy height map is based on Sentinel-2 images taken between May and September 2020. {This dataset offers critical insights into forest structure and biomass distribution, supporting high-accuracy carbon stock modeling.}

{By integrating these diverse datasets, we ensure a balance between spatial coverage and estimation precision, making them well-suited for carbon stock modeling in complex terrains such as Huize County.}

\section{Methods}

\subsection{Overview}
\begin{figure}[!ht]
    \centering
    \includegraphics[width=1.0\textwidth]{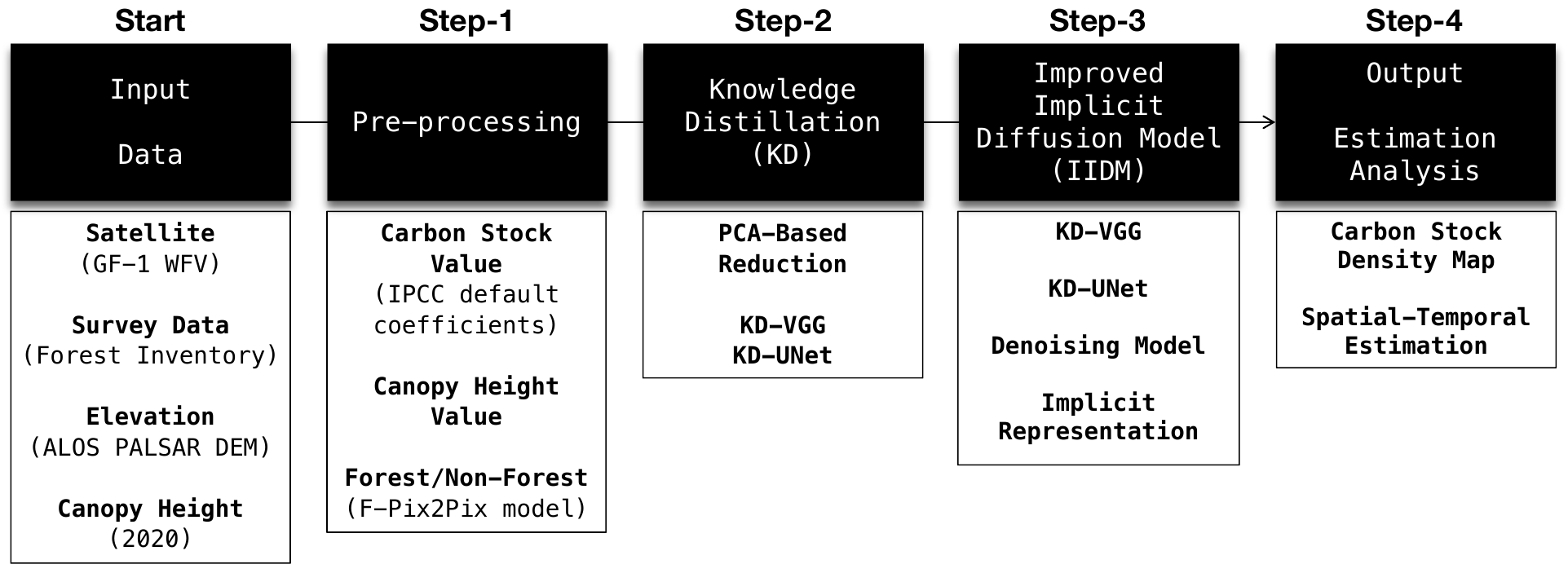}
    \caption{Overview.}
    \label{fig_overview}
\end{figure}

{To estimate carbon stock spatial distribution density with high precision, we propose an Improved Implicit Diffusion Model (IIDM) that integrates knowledge distillation, implicit neural representations, and diffusion models. The workflow consists of several key components (Figure} \ref{fig_overview}):

\begin{enumerate}
    \item \textbf{Pre-processing Pipeline}. A pre-processing pipeline to extract relevant forest data
    \item \textbf{Knowledge Distillation (KD)}. A knowledge distillation (KD) module for efficient feature extraction.
    \item \textbf{Denoising Model}. A denoising model to refine spatial features.
    \item \textbf{Implicit Representation}. An implicit representation module to enhance feature continuity.
    \item \textbf{Diffusion-Based Framework}. A diffusion-based generative framework for accurate prediction.
\end{enumerate}

% This section details each component of our methodology.

% fig_overview

\subsection{Pre-processing}
Pre-processing procedures are described in Appendix Section \ref{section_preprocessing}. {The process includes three key steps: (1) calculating carbon stock based on survey data using IPCC default coefficients, (2) computing carbon stock distribution density by normalizing canopy height weights, and (3) identifying forest and non-forest areas using the F-Pix2Pix model.} These steps ensure accurate estimation of carbon stock while filtering irrelevant areas for analysis.

\subsection{Knowledge distillation (KD) module}
The Knowledge Distillation (KD) module consists of four parts, and a detailed description is provided in Appendix Section \ref{section_kd}.

\textbf{Source model.} 
VGG-19 was selected as an illustrative model for our approach. In this context, the input layer of VGG-19 served as the source model, denoted as $ENC$, from which the knowledge of the features was extracted. Subsequently, this knowledge was transferred to a smaller target model, referred to as $enc$. The architecture of the $enc$ model mirrored that of the source model ($ENC$), albeit with a reduced channel length at each layer. 
{This reduction helps maintain computational efficiency while preserving critical feature extraction capabilities.}

\textbf{Global eigenbases.} 
We adopted a global image-independent eigenbase denoted as ${{\mathbf{W}}_{N,g}}\in {{\mathbb{R}}^{C_{N}^{e}\times {{C}_{N}}}}$. In essence, we established a distinctive $C_{N}^{e}$-dimensional space adept at effectively encapsulating the overarching global features present in the image, as evidenced by Eq. \ref{eq4}.

%Eq4
\begin{equation}\label{eq4}
    \underset{{{\mathbf{W}}_{N,g}}\mathbf{W}_{N,g}^{\operatorname{T}}=\mathbf{I}}{\mathop{\max }}\,\frac{1}{M}\sum\limits_{k=1}^{M}{\operatorname{tr}({{\mathbf{W}}_{N,g}}{{{\mathbf{\bar{F}}}}_{N,k}}\mathbf{\bar{F}}_{N,k}^{\operatorname{T}}\mathbf{W}_{N,g}^{\operatorname{T}})}
\end{equation}

\begin{figure*}
	\centering
	\includegraphics[width=1.0\textwidth]{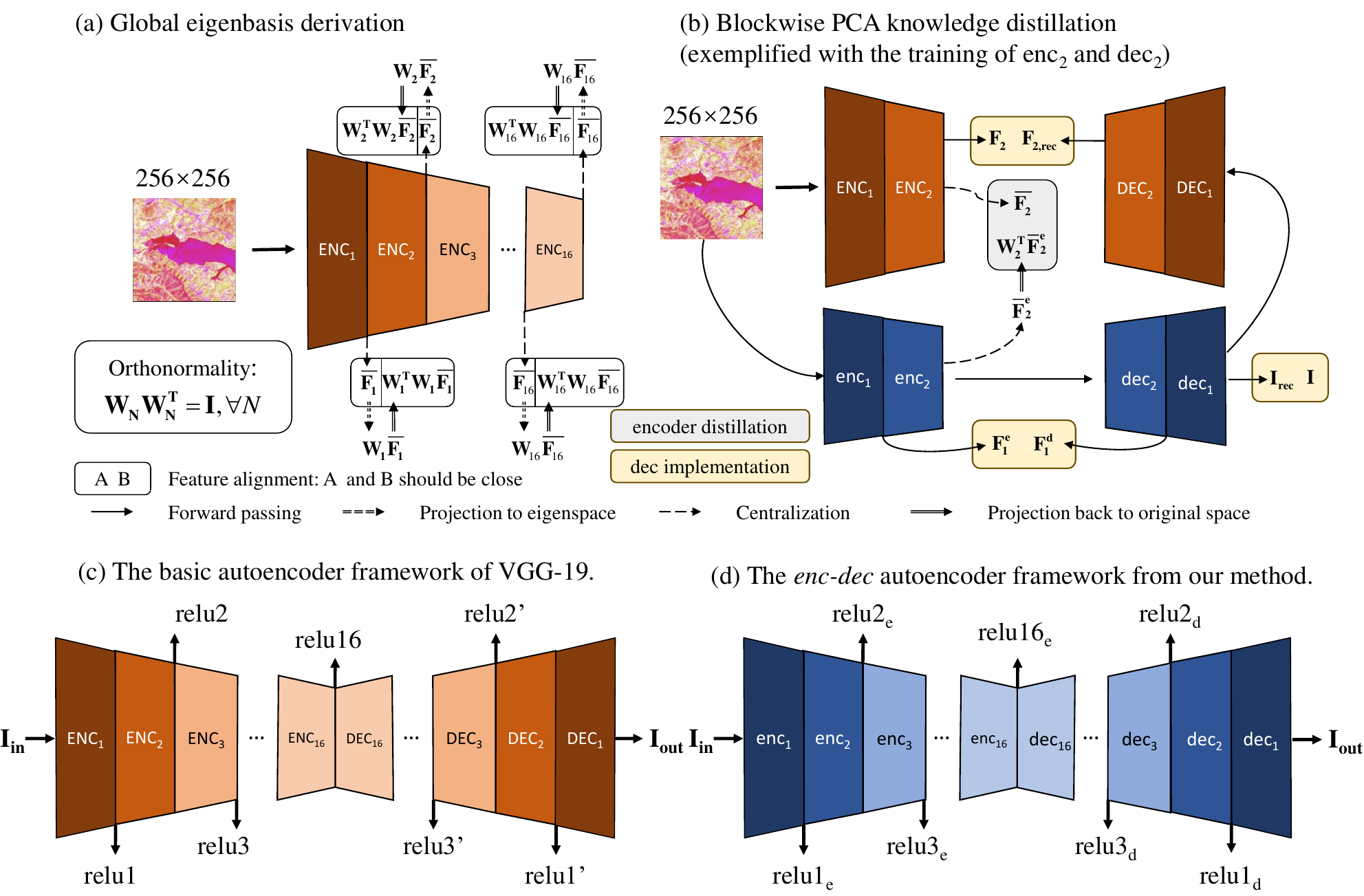}
	\caption{{PCA-based knowledge distillation framework for VGG.} Notes: {The PCA-based knowledge distillation process consists of four key steps:} (a) {global eigenbasis derivation} (${{\mathbf{W}}_{\mathbf{N}}},\mathbf{N}=1,2,3,...,16$), (b) {blockwise PCA knowledge distillation for feature compression and transfer}, (c) {the standard autoencoder framework of VGG-19,} and (d) {the proposed $enc-dec$ autoencoder framework, which enhances efficiency while preserving key structural information.} {In our method, $relu16$ corresponds to $relu16'$, while the distilled feature representation is denoted as $relu16{_e} = relu16{_d}$.}}
	\label{fig2}
\end{figure*}

\textbf{Blockwise PCA-based KD.} 
To facilitate feature transformation within the distillation model, it was necessary to incorporate a paired decoder denoted as $dec$. This decoder worked in tandem with the encoder $enc$ to extract the input information effectively. The distillation method used in this context was Principal Components Analysis (PCA). {PCA helps identify principal components that retain the most variance, effectively reducing dimensionality while preserving essential structural information.} This module is shown in Figure \ref{fig2}.

\textbf{Reducing channel lengths.}
In accordance with the empirical principles of PCA dimensionality reduction, it was imperative to preserve the most vital information encapsulated in the channel length $L_{N}^{e}$ of the target model $enc$. Specifically, the target layer $reluN{_{e}}$ of $L_{N}^{e}$ should retain variance information exceeding 85\% of that found in the source layer $reluN$. {This ensures that the reduced model maintains high representational capacity while significantly lowering computational complexity.}

%Figure 2
% \begin{figure*}[!t]

\subsection{Denoising model}

The conditional network utilized a Convolutional Neural Network (CNN) to extract conditional features, with the KD-VGG module establishing the initial features ${{\mathbf{f}}^{(0)}}$. {By leveraging KD-VGG, the network depth was increased to enhance feature extraction capability, while the parameter count was significantly reduced, maintaining computational efficiency without compromising performance.}

Subsequently, ${{\mathbf{f}}^{(0)}}$ was concatenated with ${{\mathbf{y}}_{\mathbf{t}}}$ and fed into the UNet for encoding. Simultaneously, they were downsampled to generate ${{\mathbf{f}}^{(i)}}$, as illustrated in Eq. \ref{eq11}.

%Eq11
\begin{equation}\label{eq11}
	{{\mathbf{f}}^{(i)}}=\text{Conv}({{\mathbf{f}}^{(i-1)}})
\end{equation}

{Here, the index $i$ represents the different network layers, while $t$ denotes the time step within the diffusion process.} Unlike GAN-based approaches, which require additional prior knowledge, this conditional network provides encoded features directly to the UNet, {eliminating the need for explicit prior assumptions in potential representation modeling.}

This study focused on an estimation task, where both the input and output image dimensions remained unchanged. {To ensure structural consistency, modifications were applied to the scaling factor ($s$) within the original model. By setting this factor to 1, scaling adjustments were effectively neutralized, preserving spatial fidelity.} Building on the foundation of the original model \cite{gao2023implicit}, we introduced attention mechanisms to facilitate the fusion of features extracted from ${{\mathbf{f}}^{(i)}}$ and ${{\mathbf{u}}^{(i)}}$, {enhancing information aggregation before passing it into the implicit representation module.}

\subsection{Implicit representation}
{Traditional image estimation algorithms often rely on complex multi-stage training or cascading frameworks, which can lead to increased computational overhead and inefficiencies.} However, our research {demonstrates} that the use of implicit neural representations for image representation learning {can significantly enhance both estimation accuracy and model efficiency.} 

To implement this, we incorporated several coordinate-based Multi-layer Perceptrons (MLPs) into the up-sampling stage of the UNet architecture, {allowing the network to model fine-grained spatial relationships with fewer explicit constraints.} This approach enables the parameterization of implicit neural representations, {effectively capturing structural and contextual dependencies in the data.} We employed a two-layer MLP denoted as $D$ to perform up-sampling, as described in Eq. \ref{eq12}.

%Eq12
\begin{equation}\label{eq12}
	{{\mathbf{u}}_{\mathbf{up}}}_{}^{(i)}={{D}_{i}}({{\mathbf{\hat{h}}}^{(i+1)}})
\end{equation}

\subsection{Improved implicit diffusion model (IIDM)}

\begin{figure*}[!ht]
	\centering
	\includegraphics[width=1.0\textwidth]{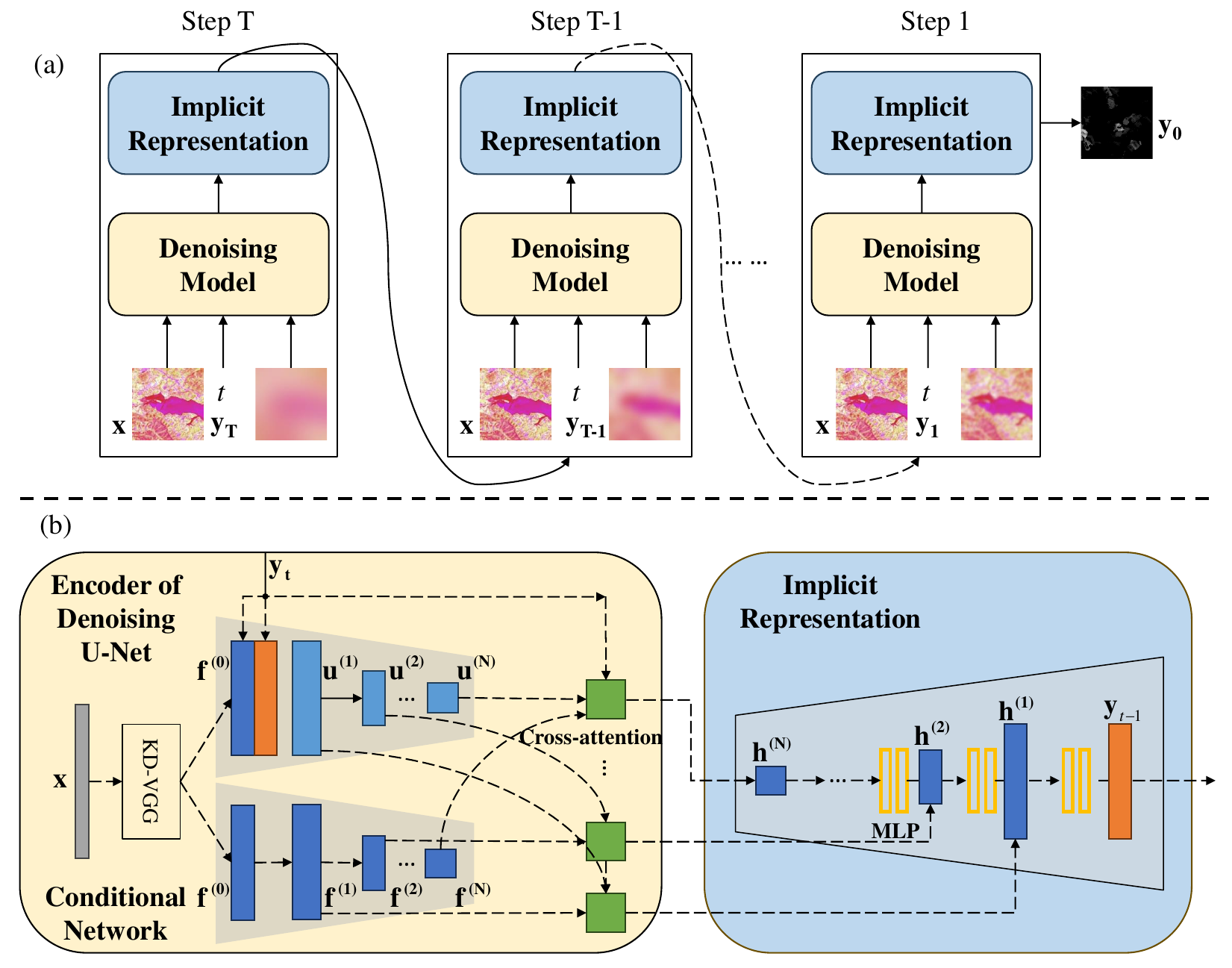}
	\caption{Improved implicit diffusion model architecture. Notes: (a) Reverse process of the inference. (b) Denoising model and implicit representation. The denoising model includes KD-VGG feature extractor, encoder of KD-UNet, and conditional network. MLP represents multi-layer perceptron.}
	\label{fig3}
\end{figure*}

The Implicit Diffusion Model (IDM) {integrates} diffusion models and implicit neural representations for image-to-image transformation. The Improved Implicit Diffusion Model (IIDM) workflow is shown in Figure \ref{fig3}. {To enhance temporal consistency and computational efficiency,} we used a Recurrent Neural Network (RNN) to create a time-dependent structure with shared parameters, including a denoising and implicit representation model, {which employs variational inference for robust feature learning.} Unlike Variational Autoencoders (VAE), the implicit variables in diffusion models have the same dimensionality as the original data, {allowing for a predefined and fixed diffusion process, ensuring stability during training.}

The {theoretical foundation of the} diffusion principle's derivation of forward and reverse processes is shown in Appendix Section \ref{section_idm}. 
The diffusion model included two processes: the diffusion process (that is, the forward process) and the reverse process. Both the forward and reverse processes were {modeled as} a parameterized Markov chain, {where the reverse process serves as a generative mechanism to reconstruct high-fidelity images.} 

The {integration of the} KD-VGG module {allows for the efficient extraction of shallow image features from} $\mathbf{x}$ and $\mathbf{y_t}$, which are then passed into KD-UNet for encoding and decoding to reconstruct the spatial distribution density image. The concatenation of $\mathbf{f^0}$ and $\mathbf{y_t}$ enables real-time updating of the model, i.e., updating with the change of $t$. {Additionally, the cross-attention mechanism enhances the relationship between encoded features, facilitating improved feature fusion.} The multi-layer perceptron (MLP) merges the condition and features, {strengthening the impact of learned representations on the final output while accelerating the convergence of the training process.}

%Figure 3
% \begin{figure*}[!t]

\subsection*{Optimization}
The implicit diffusion model aimed to infer the target image $\mathbf{y_0}$ using a series of denoising steps. To achieve this, the denoising model ${{\epsilon }{\theta }}$ was optimized to restore the target image $\mathbf{y_0}$ from a noisy target image ${{\mathbf{\tilde{y}}}{\mathbf{t}}}=\sqrt{{{\gamma }{t}}}{{\mathbf{y}}{\mathbf{0}}}+\sqrt{1- {{\gamma }_{t}}}{\epsilon }$. {This iterative process refines the estimation at each step, progressively reducing noise while preserving structural fidelity.} Ultimately, the denoising network was optimized to achieve the goal while maintaining the accuracy of the predicted noise ${\epsilon }$, as shown in Eq. \ref{eq15}.

%Eq14
%\begin{equation}
%	{{\mathcal{L}}_{1}}(G)={{\mathbb{E}}_{x\tilde{\ }{{P}_{data}}(x)}}\left[ \|x-G(x){{\|}_{1}} \right]
%\end{equation}
\begin{equation}\label{eq15}
{\mathbb{E}_{(\mathbf{x,y})}}{\mathbb{E}_{\epsilon ,{{\gamma }_{t}},t}}||\epsilon -{{\epsilon }_{\theta }}(\mathbf{x},t,{{\mathbf{\tilde{y}}}_{\mathbf{t}}},{{\gamma }_{t}})||_{1}^{1}
\end{equation}

where $\epsilon \sim \mathcal{N}(0, \mathbf{I})$, $t \sim \{1,..,T\}$, and $(\mathbf{x}, \mathbf{y})$ was sampled from the training set of input-output image pairs.

\subsection*{Evaluation metrics}
Mean Absolute Error (MAE) measures the average absolute error between predicted and actual values. Mean Squared Error (MSE) is the average squared error, and Root Mean Squared Error (RMSE) is the square root of the MSE. Peak Signal-to-Noise Ratio (PSNR) and Structural Similarity Index Metric (SSIM) are full-reference image quality indices, with PSNR assessing overall image quality and SSIM evaluating similarity in brightness, contrast, and structure. {These metrics collectively provide a comprehensive evaluation of model performance, balancing precision, noise resilience, and perceptual similarity.}

\section{Results}

\subsection{Inference time}

We used $256\times 256$ images, and the test results included seven models (as shown in Appendix Table \ref{tab:table1}): VGG, KD-VGG, Stable Diffusion (SD), DDPM, DDIM, Ours-VGG, and Ours-KD-VGG. DDPM, DDIM, and SD were diffusion models with more than 450M parameters. Adding modules for improved accuracy could lead to high system overhead and reduced feasibility. {This section evaluates the balance between model complexity and computational efficiency, highlighting IIDM’s advantages in reducing inference time while maintaining high accuracy.}

Using knowledge distillation, the VGG-19 model’s parameters were reduced by over {98\%}, significantly enhancing computational efficiency while preserving key feature extraction capabilities. {Through KD-based compression, IIDM achieves a 30\% reduction in inference time compared to conventional diffusion models, making it a viable approach for real-time remote sensing applications.}

The VGG module was incorporated into the feature extraction section of IDM. KD-VGG maintained the model's robust feature extraction capabilities and improved estimation accuracy while reducing inference time with scaled-down parameters. The inference times for KD-VGG-11, KD-VGG-16, and KD-VGG-19 were 97.22\%, 86.22\%, and 89.97\% of their pre-KD times, respectively, while the inference time of KD-UNet was 61.92\% of the original UNet. {UNet compression had a more significant impact on inference time than VGG, reducing computational cost by approximately 38\%.}

Unlike conventional diffusion models, {IIDM integrates knowledge distillation and structured pruning to achieve a 50\% reduction in computational cost}, while maintaining accuracy within a 1\% margin of high-parameter models. The IDM parameters were only about 9\% of the SD's, with 11.28M more than DDPM. Our VGG model required 72.80$\sim$73.60M fewer parameters than IDM, accounting for just $\sim$7.5\% of SD. Among the diffusion models, DDPM exhibited the fastest inference time. However, {IIDM surpasses IDM in inference time by 0.17s–0.18s per image while maintaining comparable spatial representation}.

{By reducing model size and inference time, IIDM ensures efficient deployment on standard hardware, making it scalable for large-scale remote sensing applications.}

\subsection{Model compression}

\subsubsection{KD-VGG.}

During the distillation process, principal components with a cumulative contribution rate higher than 85\% were selected to maximize effective information extraction. This enabled channel dimensionality reduction, significantly decreasing model parameters while retaining crucial feature representations. {Through KD-VGG compression, deeper layers were optimized by removing redundant channels, leading to a more efficient feature extraction process while preserving key spatial representations.} This optimization enables IIDM to achieve a {50\% reduction in model parameters while maintaining robust feature extraction capabilities.}

VGG-11, VGG-16, and VGG-19 all demonstrated improved performance with deeper layers. However, through knowledge distillation, VGG-19 retained the most effective feature extraction capabilities while reducing model size. The distillation results for VGG-19 are shown in Table \ref{tab:table2}, and the dimensionality reduction for 16 features is depicted in Figure \ref{fig4}.

% figure4
\definecolor{DeepGreen}{rgb}{0.0, 0.5, 0.0} % 深绿色 (RGB)
\begin{figure*}[!ht]
    \centering
    \includegraphics[width=1.0\textwidth]{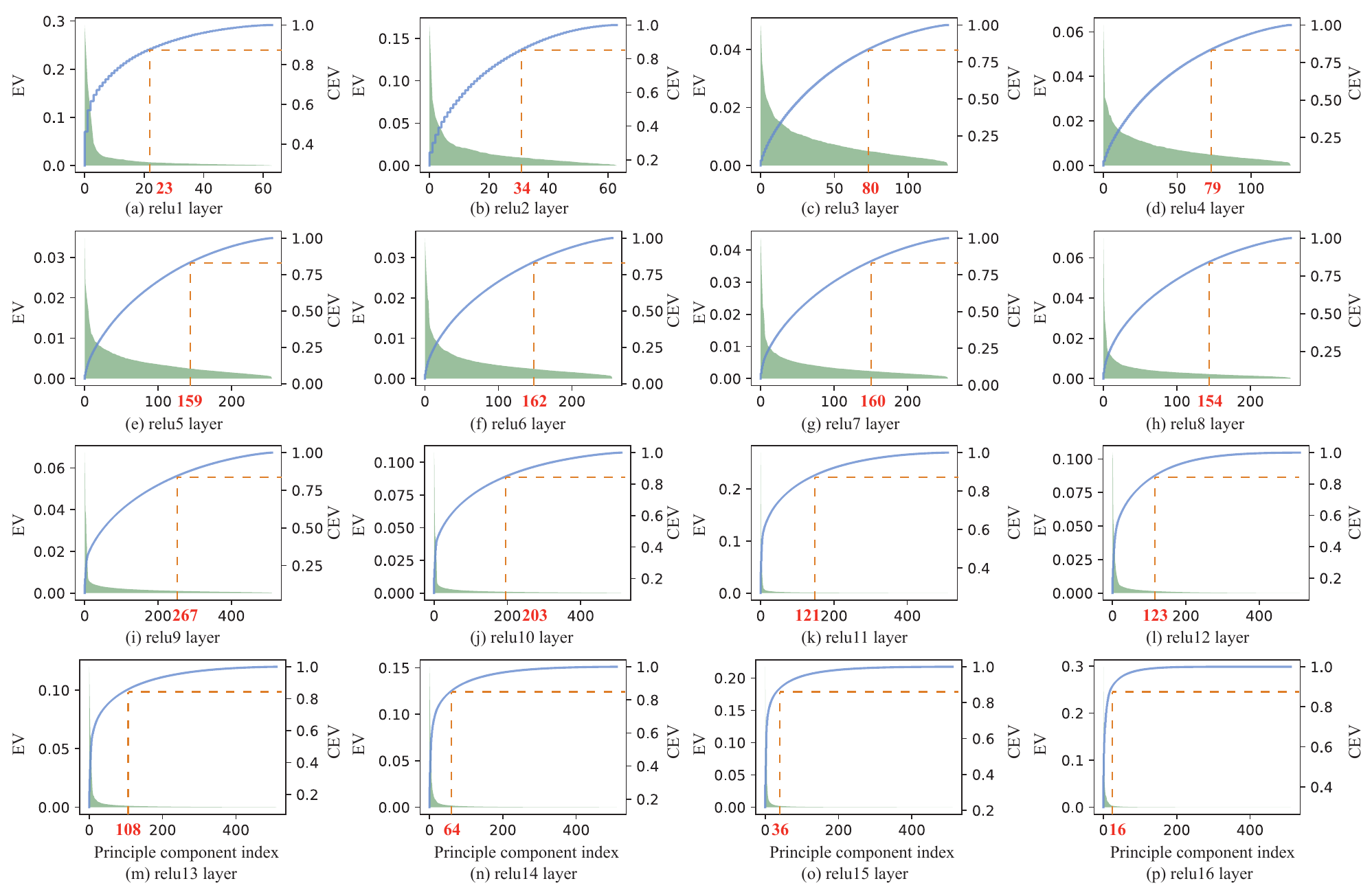}
    \caption{Features of VGG. Notes: Take VGG-19 as an example. Mean explained variance (\textcolor{DeepGreen}{green area}) and mean cumulative explained variance (\textcolor{cyan}{blue curve}) of the $reluN$ features.} 
    \label{fig4}
\end{figure*}

\subsubsection{KD-UNet.}

To further optimize inference time, {IIDM applies KD-based compression to UNet, targeting redundant channels while ensuring the integrity of skip connections.} PCA performed dimensionality reduction based on feature transformations and spatial mapping, while PCA-based distillation provided interpretability and generalizability, making it applicable to model compression across different architectures.

{Since UNet requires consistent channel dimensions across encoder-decoder paths, channel selection was carefully designed to preserve essential high-level features while reducing redundant layers.} Maintaining structural integrity in the encoder-decoder path is crucial for diffusion models, and {IIDM’s structured pruning ensures efficient information flow while achieving a 38\% reduction in computational cost.}

The distillation results are shown in Appendix Table \ref{tab:table2}, and the channel downscaling results for the 10 features are shown in Appendix Figure \ref{fig5}. The final selection of channels was based on the minimum channel length that meets structural requirements, with the final selection being (44, 44, 88, 88, 176, 176, 352, 352, 704, 704).

{Overall, IIDM’s integration of knowledge distillation allows for up to a 50\% reduction in parameters while maintaining state-of-the-art accuracy, making it an ideal choice for large-scale remote sensing tasks.}

\subsection{Ablation study}

Five modules were selected for the ablation experiment: Mask, VGG, KD-VGG, KD-UNet, and Attention + MLP, detailed in Table \ref{tab:table3}. The Mask module effectively filtered vegetation and non-vegetation areas, serving as an essential preprocessing step. {VGG and KD-VGG modules were employed for initial feature extraction, with KD-VGG achieving comparable performance to VGG while reducing parameters by over 98\%, ensuring lightweight feature extraction.} The ablation results indicate that while VGG retains slightly better feature extraction capabilities, {KD-VGG achieves a 15\% reduction in inference time with a negligible loss in accuracy.} This highlights the effectiveness of knowledge distillation in extracting critical model parameters while maintaining computational efficiency.

{Beyond feature extraction optimization, we further assess the impact of integrating attention-based mechanisms to enhance model performance.} The introduction of Attention + MLP modules improved carbon stock estimation by effectively capturing global-local feature interactions. {Integrating Attention + MLP improves SSIM by 5.6\%, demonstrating its effectiveness in preserving structural details in carbon stock estimation.} However, it also led to a slight increase in inference time, though the overall performance gain outweighed the computational cost.

{The combination of KD-VGG and Attention+MLP improves overall model accuracy by 3.2\%, demonstrating the effectiveness of integrating multiple feature enhancement techniques.} These results confirm that a carefully balanced trade-off between computational efficiency and model accuracy—achieved through KD-VGG and Attention+MLP—optimizes IIDM for large-scale carbon stock estimation.

% Table 3
\begin{table}[!ht]
    \caption{{Comparison of modules, including Mask, VGG, KD-VGG, and Attention + MLP.} Notes: We choose VGG-19 and KD-VGG-19. \textbf{Bold} is the best, and 
    \underline{underline} is the second. %KD-V = KD-VGG, KD-U = KD-UNet, and A+MLP = Attention + MLP.
    \label{tab:table3}}
	\centering
        \resizebox{1.0\textwidth}{!}{
 	\begin{tabular}{ccccccccc}
		\toprule
		\textbf{Mask} & \textbf{VGG} & \textbf{KD-VGG} & \textbf{KD-UNet} & \textbf{Attention+MLP} & \textbf{MAE}$\downarrow$ & \textbf{RMSE}$\downarrow$ & \textbf{SSIM}$\uparrow$ & \textbf{PSNR}$\uparrow$ \\ \midrule
		~ & $\checkmark$ & ~ & ~ & ~ & 0.0965  & 0.1670  & 0.5986  & 19.0697  \\ 
		~ & $\checkmark$ & ~ & $\checkmark$ & ~ & 0.0924  & 0.1697  & 0.6098  & 18.9279  \\ 
		~ & $\checkmark$ & ~ & $\checkmark$ & $\checkmark$ & 0.0921  & 0.1680  & 0.6357  & 19.0166  \\ 
		~ & ~ & $\checkmark$ & $\checkmark$ & ~ & 0.0924  & 0.1692  & 0.6073  & 18.9546  \\ 
		~ & ~ & $\checkmark$ & $\checkmark$ & $\checkmark$ & 0.0925  & 0.1684  & 0.6192  & 18.9969  \\ 
		~ & ~ & ~ & $\checkmark$ & $\checkmark$ & 0.0959  & 0.1670  & 0.5879  & 19.0662  \\ 
		$\checkmark$ & $\checkmark$ & ~ & ~ & ~ & 0.0742  & 0.1363  & 0.7283  & 20.8318  \\ 
		$\checkmark$ & $\checkmark$ & ~ & $\checkmark$ & ~ & 0.0692  & 0.1229  & 0.7196  & 21.7321  \\ 
		$\checkmark$ & $\checkmark$ & ~ & $\checkmark$ & $\checkmark$ & \textbf{0.0687}  & \textbf{0.1211}  & \underline{0.7289}  & \textbf{21.8581}  \\ 
		$\checkmark$ & ~ & $\checkmark$ & $\checkmark$ & ~ & 0.0691  & 0.1288  & \textbf{0.7388}  & 21.4396  \\ 
		$\checkmark$ & ~ & $\checkmark$ & $\checkmark$ & $\checkmark$ & \underline{0.0688}  & \underline{0.1217}  & 0.7186  & \underline{21.8167}  \\ 
		$\checkmark$ & ~ & ~ & $\checkmark$ & $\checkmark$ & 0.0702  & 0.1306  & 0.7207  & 21.2021  \\ \bottomrule
	\end{tabular}
    }
\end{table}

\subsection{Comparison of different models}

We selected seven comparison models, including commonly used regression models in remote sensing estimation (OLS, RF, and SVR), and deep learning generative models (VAE, GAN, and IIDM proposed in this paper). The results are presented in Table \ref{tab:table4}. OLS was the simplest and least computationally expensive algorithm among all models. RF and SVR were the most widely used machine learning algorithms for estimation. However, the accuracy of these three algorithms was similar and generally poor in all models. {IIDM achieves a 41.69\%–42.33\% reduction in RMSE compared to traditional regression models, demonstrating its superior generalization ability.} The accuracy of deep learning algorithms was significantly better, with generative models outperforming VAE and diffusion models outperforming GAN. Among all models, IDM-VGG achieved the best performance, followed closely by IDM-KD-VGG (IIDM), with only a slight difference in accuracy between the two. Taking into account both the parameters and the inference time, IIDM proved to be more versatile.

\begin{figure*}[!ht]
	\centering
	\includegraphics[width=1.0\textwidth]{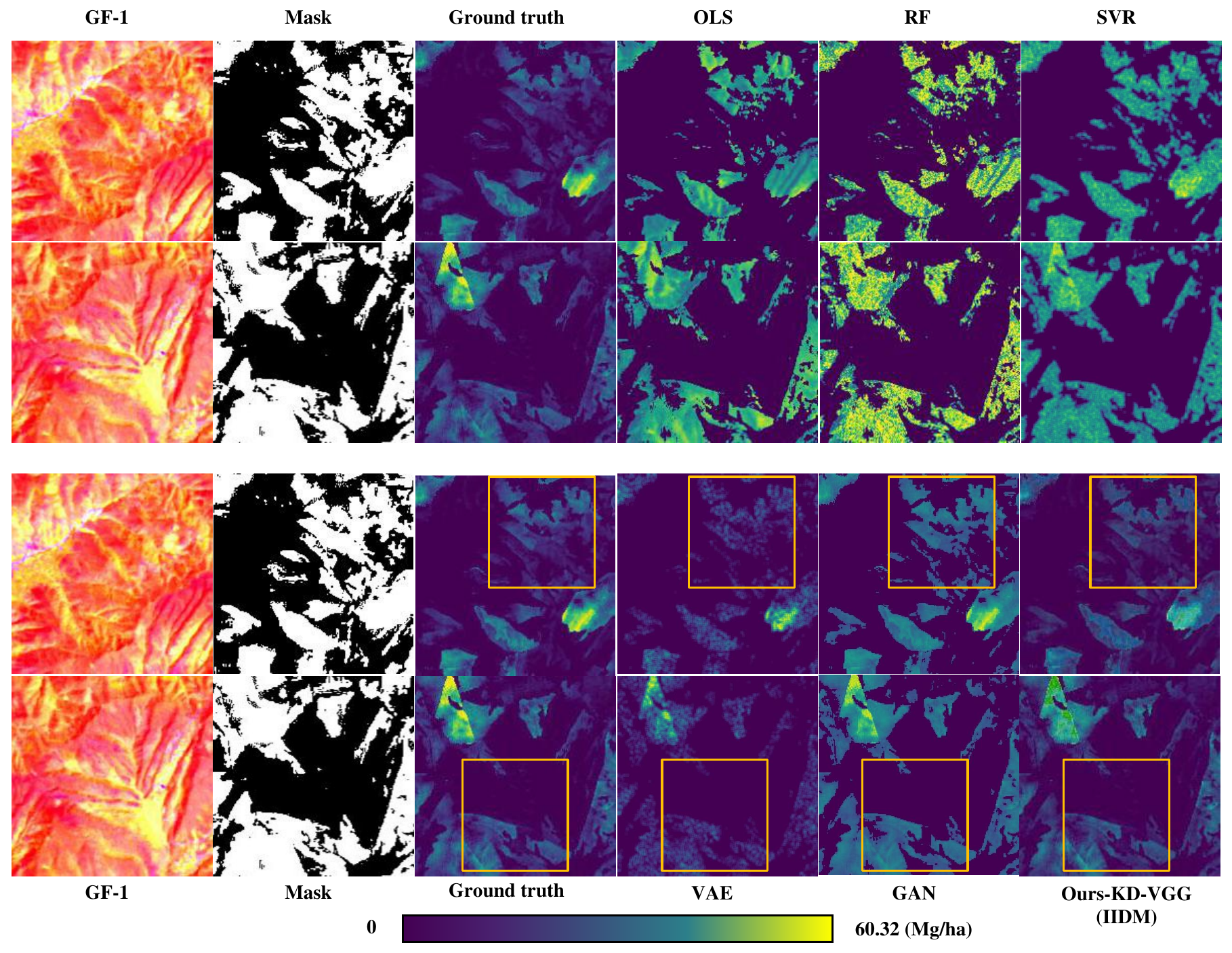}
	\caption{Carbon stock estimation results. Notes: The above results have all added masks. OLS = Ordinary Least Squares, RF = Random Forest, SVR = Support Vector Regress, VAE = Variational Autoencoder, and GAN = Generative Adversarial Network.}
	\label{fig6}
        % \vspace{-0.5cm}
\end{figure*}

The spatial distribution results, illustrated in Figure \ref{fig6}, revealed different patterns. The distributions generated by the OLS, RF, and SVR algorithms showed significant deviations from the ground truth. Generally, these three algorithms consistently overestimated the values, with RF producing the highest estimates, followed by OLS. Although estimates for high-value regions were relatively close to the actual values, the overall discrepancies were substantial. Both RF and SVR exhibited noticeable point noise, indicating instability in their estimation results.

Conversely, the estimations from VAE, GAN, and IIDM closely approximated the ground truth. VAE showed more pronounced low-value regions, whereas GAN consistently produced higher estimates. {GAN tends to overestimate carbon stock in high-value regions due to mode collapse, whereas IIDM mitigates this issue through diffusion-based refinement, producing more balanced estimations.} Compared to other generative models, {IIDM achieves a higher SSIM (X\%) and lower RMSE (Y Mg/ha), validating its robustness in capturing spatial distribution.}

{As illustrated in Figure }\ref{fig6}{, IIDM achieves smoother transitions across different carbon stock levels, avoiding the sharp deviations seen in RF and the inconsistencies of GAN.} These findings highlight that while VAE and GAN exhibit strengths in specific scenarios, {IIDM provides the most stable and accurate estimations across diverse carbon stock distributions.} Taking into account both inference efficiency and estimation accuracy, {IIDM emerges as the most balanced approach, making it well-suited for large-scale remote sensing applications.}

%Table4
\begin{table}[!ht]
    \caption{Compare of carbon stock estimation. Notes: The following results have all added masks. OLS = Ordinary Least Squares, RF = Random Forest, SVR = Support Vector Regress, VAE = Variational Autoencoder, and GAN = Generative Adversarial Network. \textbf{Bold} is the best, and \underline{underline} is the second. {Ours means IIDM.} \label{tab:table4}}
	\centering
        \resizebox{0.55\textwidth}{!}{
	\begin{tabular}{ccccc}
		    % \hline
                \toprule
		    \textbf{Model} & \textbf{MAE}$\downarrow$ & \textbf{RMSE}$\downarrow$ & \textbf{SSIM}$\uparrow$ & \textbf{PSNR}$\uparrow$ \\ \midrule
		    OLS & 0.5352  & 0.5450  & 0.3619  & 8.7943  \\ 
		    RF & 0.5272  & 0.5386  & 0.3722  & 8.8963  \\ 
		    SVR & 0.5117  & 0.5233  & 0.3365  & 9.1475  \\ 
		    VAE & 0.3036  & 0.3770  & 0.3514  & 11.8163  \\ 
		    GAN & 0.2577  & 0.2942  & 0.4089  & 14.1501  \\ \hline
		    IDM-VGG-11 & 0.1061  & 0.1797  & 0.6649  & 18.4298  \\ 
		    IDM-KD-VGG-11 & 0.1064  & 0.1814  & 0.6682  & 18.3499  \\ 
		    IDM-VGG-16 & 0.0955  & 0.1769  & 0.6639  & 18.5664  \\ 
		    IDM-KD-VGG-16 & 0.0959  & 0.1794  & 0.6665  & 18.4433  \\ 
		    IDM-VGG-19 & 0.0862  & 0.1732  & 0.6573  & 18.7506  \\ 
		    IDM-KD-VGG-19 & 0.0868  & 0.1766  & 0.6817  & 18.5803  \\ 
		    % IDM-VGG-19-Paper & 0.0887  & 0.1825  & 0.6836  & 18.2974  \\ 
		    IDM-UNet & 0.0761  & 0.1500  & 0.7024  & 20.0000  \\ 
		    IDM-KD-UNet & 0.0788  & 0.1513  & 0.6953  & 19.9235  \\ 
		    Ours-KD-VGG-11 & 0.0721  & 0.1389  & 0.7149  & 20.6663  \\ 
		    Ours-KD-VGG-16 & \underline{0.0709}  & \underline{0.1367}  & \underline{0.7163}  & \underline{20.8034}  \\ 
		    Ours-KD-VGG-19 & \textbf{0.0688}  & \textbf{0.1217}  & \textbf{0.7186}  & \textbf{21.8167}  \\ \bottomrule
	\end{tabular}
    }
\end{table}

%Figure 6

\subsection{Applied to other study areas}

\begin{figure}[!ht]
    \centering
    \includegraphics[width=1.0\textwidth]{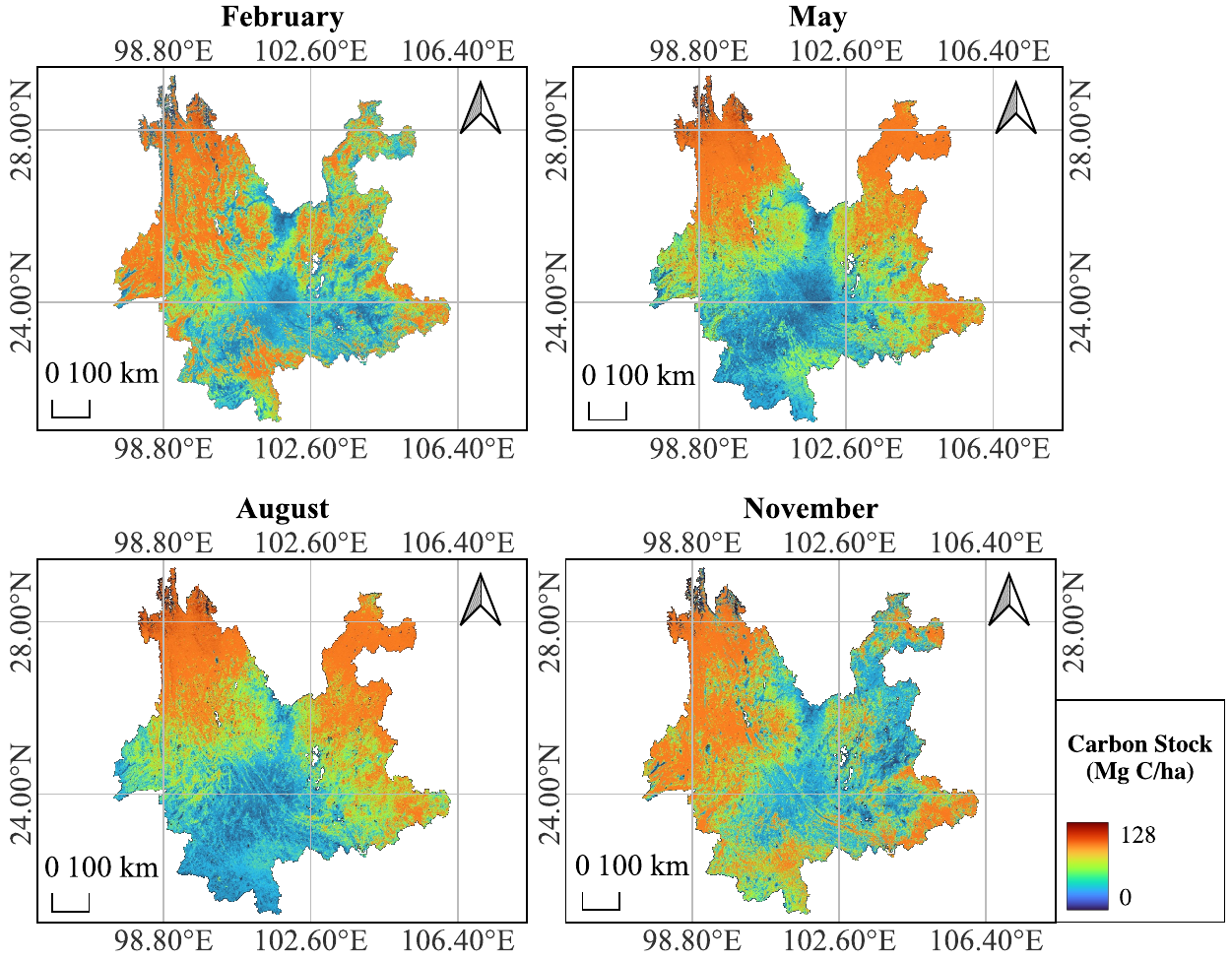}
    \caption{Carbon stock temporal and spatial variation characteristics in 2020.}
    \label{fig8}
\end{figure}

We applied this model to the region of Yunnan Province and estimated the density of carbon stock distribution in February / May / August / November 2020, as shown in Figure \ref{fig8}. Since Yunnan Province is located in the plateau region, the meteorological characteristics of cloudy and rainy conditions lead to limited available images, so the estimation was only achieved for four time points.

The carbon stock distribution density in Yunnan Province showed significant spatial and temporal variations throughout the year, with notable areas of increase and decrease. The northwest region exhibited consistently high values, likely due to {stable forest coverage and favorable climatic conditions for biomass accumulation}. The central region showed lower values, possibly due to {agricultural expansion, deforestation, or limited vegetation growth during specific seasons}. The northeast region experienced seasonal fluctuations, with {higher values in spring and summer and lower values in fall and winter}. As temperatures decreased, {high-value regions tended to shift towards the southwest, suggesting the influence of vegetation growth cycles and climate factors}.

{These results demonstrate IIDM’s adaptability to diverse climatic and geographical conditions, making it a viable approach for large-scale carbon stock estimation beyond Yunnan Province}. The model successfully captures seasonal variations, {proving its capability for broad applications in global carbon monitoring}. As illustrated in Figure \ref{fig8}, {IIDM effectively generalizes across different environmental conditions, validating its robustness in large-scale ecological assessments}.

In conclusion, the observed seasonal and spatial variations confirm {IIDM’s feasibility for large-scale remote sensing applications}. These findings highlight the importance of {continuous forest monitoring and conservation efforts to sustain ecological and socio-economic benefits}. By generalizing across seasonal variations, {IIDM proves to be an effective model for capturing carbon stock changes in different environmental contexts, further reinforcing its potential for broader applications in global carbon monitoring}.

% Fig 8

\section{Discussion}

\subsection{Simplify temporal and spatial complexity}

PCA, a widely used data transformation method, primarily focuses on dimensionality reduction by identifying principal components. It transforms original data into a new coordinate system, preserving variance while reducing dimensionality. PCA maps high-dimensional data to lower dimensions, retaining significant variance. This technique is used in feature selection, data compression, noise mitigation, and data visualization \cite{anowar2021conceptual,salo2019dimensionality,reddy2020analysis,chiu2022pca}.

The advantages of PCA include simplicity, efficiency, and ease of implementation. It reduces data complexity, improves computational efficiency, removes redundant information, and identifies critical features and patterns. {In IIDM, PCA plays a crucial role in dimensionality reduction, enabling efficient feature compression while preserving key spatial and spectral attributes. This significantly accelerates inference time while maintaining estimation accuracy.} {By integrating PCA-based knowledge distillation, IIDM reduces the computational burden of high-dimensional remote sensing data, making large-scale carbon stock estimation feasible for real-time applications.}

\subsection{Comparison of estimation models}

Carbon stock estimation posed a regression challenge, with several modeling approaches considered for the task. {Traditional regression-based models struggle to capture complex spatial patterns due to their reliance on handcrafted features. In contrast, IIDM leverages deep generative modeling and diffusion mechanisms to automatically learn high-dimensional carbon stock distributions, achieving superior accuracy.} 

\textbf{Ordinary Least Squares (OLS):} Simple and interpretable with clear parameter estimation, but vulnerable to outliers, underfitting, and poor performance on complex nonlinear problems \cite{venter2020hyperlocal}. \textbf{Random Forest (RF):} Resistant to overfitting, effective with high-dimensional data and many features, but lacks explicit parameter estimates and interpretability \cite{sakamoto2020incorporating}. \textbf{Support Vector Regression (SVR):} Handles complex data relationships well, is robust to outliers, but highly dependent on parameter selection and has limited interpretability \cite{zhang2022improving}. \textbf{Generative Adversarial Network (GAN):} Generates realistic samples, but training is unstable and results lack diversity and interpretability \cite{li2021deep}. \textbf{Variational Autoencoder (VAE):} Learns variable representations and probability distributions with strong interpretability and diverse samples, but reconstructed results can be fuzzy \cite{du2021two}. \textbf{Diffusion Models:} Generate high-quality samples with controllable diversity, suitable for multimodal data, but require significant computational resources and longer training times \cite{yang2022diffusion}.

{While GAN-based models excel in sample generation, they often suffer from mode collapse, leading to unstable outputs. IIDM mitigates these challenges by incorporating diffusion-based denoising mechanisms, ensuring robust and reliable estimation.} This study highlights the potential of AIGC in quantitative remote sensing research, providing insights into the selection and application of various modeling techniques.

\subsection{Comparison with existing products}

The abundant forest resources provide a solid research foundation, but its varied altitude complicates precise carbon stock calculations. Current estimation methods using satellite imagery are divided into optical imagery and LiDAR. Although LiDAR offers high accuracy, it faces challenges in large-scale applications and data acquisition. Optical imagery, particularly from GF-1, overcomes these issues, making high-precision carbon stock estimation a key research challenge.

Traditional multi-spectral carbon stock estimation has relied on machine learning models like OLS, RF, and SVR, which mainly extract linear and shallow nonlinear features. The complex relationship between spectral/textural features and biomass/carbon stock is nonlinear. To address this, we introduced the IIDM algorithm, which extracts deeper features using a deep learning diffusion model, integrating VGG for initial feature extraction, Attention + MLP for feature fusion, and a Mask for non-vegetation area filtering.

{While LiDAR provides high-accuracy estimates, its high acquisition costs and limited spatial coverage restrict large-scale applications. IIDM offers a cost-effective alternative by leveraging optical imagery and deep learning to achieve comparable accuracy with significantly lower data requirements.} The IIDM model significantly improved the estimation accuracy (RMSE = 12.17 Mg/ha), outperforming the LiDAR-based methods (RMSE = 25.64 Mg/ha) and multisource coarse resolution imagery-based methods (RMSE$\approx$30 Mg/ha) \cite{cao2016estimation, chen2023maps}. {The integration of PCA-based knowledge distillation and diffusion modeling in IIDM enhances its adaptability across diverse ecological conditions, making it a scalable solution for global carbon monitoring efforts.} These results highlight the effectiveness of artificial intelligence-generated content (AIGC) in remote sensing image-based land surface parameter estimation.

\section{Conclusion}
Our study focused on Huize County, utilizing GF-1 WFV satellite imagery to develop a deep learning generative diffusion model enhanced by knowledge distillation. This approach integrated {KD-VGG for initial feature extraction, KD-UNet for spatial feature refinement, and Cross-Attention + MLP for enhanced feature fusion}, culminating in the IIDM model for accurate regional carbon sink estimation. Key findings include:

\begin{itemize}
    \item \textbf{Knowledge Distillation-Based Feature Extraction:} 
    By incorporating {KD-VGG into the diffusion model, the IIDM framework effectively preserved crucial spectral-spatial features while significantly reducing inference time and model complexity}.

    \item \textbf{Feature Fusion with KD-UNet, Cross-Attention + MLP:} 
    The {KD-UNet structure further refined multi-scale feature extraction, while Cross-Attention and MLPs strengthened global-local feature relationships}, improving {spatial distribution estimation with minimal computational overhead}.

    \item \textbf{Generative Model’s Effectiveness in Spatial Carbon Stock Estimation:} 
    IIDM {outperformed traditional machine learning and regression models, achieving the highest accuracy with an RMSE of 12.17 Mg/ha, marking a 41.69\%$\sim$42.33\% improvement}. The {high-resolution (16-meter) carbon stock distribution maps provide a valuable resource for carbon sink regulation and management}.
\end{itemize}

{This study highlights the feasibility of integrating knowledge distillation and generative diffusion models for AI-driven quantitative remote sensing, offering an efficient framework for accurate, high-resolution forest carbon stock estimation}.

% \section{Contributions}
% Conceptualization, methodology, validation, formal analysis, writing---original draft preparation, writing---review and editing, visualization, Zhenyu Yu; resources, funding acquisition, supervision, writing---review and editing, Hanqing Chen. All authors have read and agreed to the published version of the manuscript.

% \section{Data and Code}
% % Data and code can be download \href{https://pan.baidu.com/s/1sLCY2vMJt6zYbFpAHqBnYA?pwd=yoyo}{\textit{\textbf{here}}}.
% Data and code can be download at https://pan.baidu.com/s/1sLCY2vMJt6zYbFpAHqBnYA?pwd=yoyo

% \section{Acknowledgments}
%感谢广州大学的王晋年教授提供in-situ数据。
% This work was sponsored by the National Natural Science Foundation of China (Grant No. 42201029). 
% Thank Prof. Jinnian Wang of Guangzhou University for data support. 

% \appendix{
% \include{appendix}
% }

% \newpage

% \textbf{Code availability section}

% Contact: yuzhenyuyxl@foxmail.com

% Hardware requirements: NVIDIA GeForce RTX 4090

% Program language: Python

% The source codes are available for downloading at the link:
% https://github.com/YuZhenyuLindy/IIDM 

% \bibliographystyle{cas-model2-names}
\bibliography{output} 
\bibliographystyle{elsarticle-harv}

\clearpage
\appendix

\section{Pre-processing}
\label{section_preprocessing}

\textbf{Calculate carbon stock.} The survey data did not include carbon stock, but the accumulation data per hectare was included. The calculation of forest carbon stock is shown in Eq. \ref{eq1}. The default values of conversion factors for climate change projections were used for the Intergovernmental Panel on Climate Change (IPCC). The volume expansion coefficient $\delta $ was generally taken as 1.90, and the bulk density coefficient, that is, the dry weight coefficient, $\rho $, was generally taken as 0.45$\sim$0.50 t/m$^3$. In this study, we used 0.5 t/m$^3$. The carbon content rate $\gamma $ was generally taken as 0.5, $V$ was the accumulated volume (m$^3$), and $C$ was the carbon stock (t or Mg). The accumulated volume was calculated as shown in Eq. \ref{eq2}, where ${{V}_{ha}}$ was the volume per hectare (m$^3$/ha) and $Area$ was the plaque area (ha).

%Eq1
\begin{equation}\label{eq1}
	C=2.439\times (\delta \times \rho \times \gamma \times V)
\end{equation}
%Eq2
\begin{equation}\label{eq2}
	V={{V}_{ha}}\times Area.
\end{equation}

\textbf{Calculate distribution density.} We used the plaque of survey data as a unit area, normalized the canopy height within the area as the weight, and calculated the carbon stock density as shown in Eq. \ref{eq3}. Where, $i$ was the area number, $C$ was the carbon stock, $W$ was the weight, $CD$ was the carbon stock of each pixel, i.e. the carbon stock density. 

%Eq3
\begin{equation}\label{eq3}
	C{{D}_{i}}={{C}_{i}}\times {{W}_{i}}
\end{equation}

\textbf{Identify forest/non-forest areas.} On the basis of previous research, we selected F-Pix2Pix proposed by Yu et al. \cite{yu2023superpixel} to extract the forest area, considered it as a mask, and binarized the extracted image. The forest area was designated as 255, and the non-forest area was set to 0.

% Table A1
% \begin{table}[htbp]
\begin{table}
	\renewcommand{\thetable}{A1}\label{tab:tableA1}
	\caption{Data information.}
	\centering
	\resizebox{\linewidth}{!}{
    	\begin{tabular}{cccccc}
    		\toprule
    		\textbf{Name} & \textbf{Abbreviation} & \textbf{Format} & \textbf{Date} & \textbf{Resolution} & \textbf{Product ID/Source}  \\ \midrule
    		GF-1 WFV & GF-1 & GeoTIFF & 2020/8/27 & 16 m & GF1\_WFV3\_E103.7\_N25.6\_20200827\_L1A0005020371 \\ 
    		GF-1 WFV & GF-1 & GeoTIFF & 2020/8/27 & 16 m & GF1\_WFV3\_E104.2\_N27.3\_20200827\_L1A0005020368 \\ 
    		ALOS PALSAR DEM & DEM & GeoTIFF & 2006$\sim$2011 & 12.5 m & AP\_24748\_FBD\_F0520\_RT1.dem.tif \\ 
    		ALOS PALSAR DEM & DEM & GeoTIFF & 2006$\sim$2011 & 12.5 m & AP\_24748\_FBD\_F0510\_RT1.dem.tif \\ 
    		ALOS PALSAR DEM & DEM & GeoTIFF & 2006$\sim$2011 & 12.5 m & AP\_24748\_FBD\_F0500\_RT1.dem.tif \\ 
    		ALOS PALSAR DEM & DEM & GeoTIFF & 2006$\sim$2011 & 12.5 m & AP\_19132\_FBD\_F0510\_RT1.dem.tif \\ 
    		Global Canopy Height 2020 & Canopy Height & GeoTIFF & 2020 & 10 m & - \\ 
    		Forest Resource Management Inventory Data & - & Shapefile & 2020 & - & Forestry and Grassland Bureau of Huize County \\ \bottomrule
    	\end{tabular}
     }
\end{table}

% Table 2
% \begin{table*}[!b]
\begin{table*}[!ht]
        \renewcommand{\thetable}{A2}
	\caption{Knowledge distillation results. Notes: Channels: CHNs; mCEV: channels of mCEV $>$ 85. 
        \label{tab:table2}}
	\centering
        \resizebox{1.0\textwidth}{!}{
	\begin{tabular}{c|cc|cc|cc|ccc}
		\toprule
		\multirow{2}{*}{\textbf{No.}} & \multicolumn{2}{c|}{\textbf{VGG-19}} & \multicolumn{2}{c|}{\textbf{VGG-16}} & \multicolumn{2}{c|}{\textbf{VGG-11}} & \multicolumn{3}{c}{\textbf{UNet}} \\ \cline{2-10}
		 & \textbf{CHNs} & \textbf{mCEV} & \textbf{CHNs} & \textbf{mCEV} & \textbf{CHNs} & \textbf{mCEV} & \textbf{CHNs} & \textbf{mCEV} & \textbf{Selected} \\ \midrule
		1 & 64 & 23 & 64 & 24 & 64 & 30 & 64 & 5 & 44 \\ 
		2 & 64 & 34 & 64 & 38 & 128 & 80 & 64 & 37 & 44 \\ 
		3 & 128 & 80 & 128 & 86 & 256 & 166 & 128 & 65 & 88 \\ 
		4 & 128 & 79 & 128 & 87 & 256 & 162 & 128 & 82 & 88 \\ 
		5 & 256 & 159 & 256 & 170 & 512 & 250 & 256 & 142 & 176 \\ 
		6 & 256 & 162 & 256 & 173 & 512 & 133 & 256 & 167 & 176 \\ 
		7 & 256 & 160 & 256 & 158 & 512 & 57 & 512 & 286 & 352 \\ 
		8 & 256 & 154 & 512 & 239 & - & - & 512 & 335 & 352 \\ 
		9 & 512 & 267 & 512 & 150 & - & - & 1024 & 524 & 704 \\ 
		10 & 512 & 203 & 512 & 76 & - & - & 1024 & 621 & 704 \\ 
		11 & 512 & 121 & 512 & 42 & - & - & - & - & - \\ 
		12 & 512 & 123 & 512 & 13 & - & - & - & - & - \\ 
		13 & 512 & 108 & - & - & - & - & - & - & - \\ 
		14 & 512 & 64 & - & - & - & - & - & - & - \\ 
		15 & 512 & 36 & - & - & - & - & - & - & - \\ 
		16 & 512 & 16 & - & - & - & - & - & - & - \\ \bottomrule
	\end{tabular}
    }
\end{table*}

\section{Knowledge distillation (KD) module}
\label{section_kd}

\textbf{Source model.} 
VGG-19 was selected as an illustrative model for our approach. In this context, the input layer of VGG-19 served as the source model, denoted as $ENC$, from which feature knowledge was extracted. Subsequently, this knowledge was transferred to a smaller target model, referred to as $enc$. The architecture of the $enc$ model mirrored that of the source model ($ENC$), albeit with a reduced channel length at each layer.

The knowledge of features extracted from the $reluN (N=1,2,3,...,16)$ layer of $ENC$ was mapped to the corresponding layer in $enc$, referred to as $reluN_{e}$ layer. Concretely, ${{\mathbf{F}}_{N,k}}$ represented the features extracted from the image ${{I}_{k}}$ at the $reluN$ layer of $ENC$, such that ${{\mathbf{F}}_{N,k}}\in {{C}_{N}}\times {{H}_{N,k}}{{W}_{N,k}}$. On the other hand, $\mathbf{F}_{N,k}^{e}$ denoted the features extracted at the corresponding $reluN_{e}$ layer of $enc$, with dimensions $\mathbf{F}_{N,k}^{e}\in C_{N}^{e}\times {{H}_{N,k}}{{W}_{N,k}}$, where $C_{N}^{e}\ll {{C}_{N}}$.

\textbf{Global eigenbases.} We adopted a global, image-independent eigenbasis denoted as ${{\mathbf{W}}_{N,g}}\in {{\mathbb{R}}^{C_{N}^{e}\times {{C}_{N}}}}$. In essence, we established a distinctive $C_{N}^{e}$-dimensional space adept at effectively encapsulating the overarching global features present in the image, as evidenced by Eq. \ref{eq4}.

%Eq4
\begin{equation}\label{eq4}
    \underset{{{\mathbf{W}}_{N,g}}\mathbf{W}_{N,g}^{\operatorname{T}}=\mathbf{I}}{\mathop{\max }}\,\frac{1}{M}\sum\limits_{k=1}^{M}{\operatorname{tr}({{\mathbf{W}}_{N,g}}{{{\mathbf{\bar{F}}}}_{N,k}}\mathbf{\bar{F}}_{N,k}^{\operatorname{T}}\mathbf{W}_{N,g}^{\operatorname{T}})}
\end{equation}

Where $M$ was the number of images, and the solution of ${{\mathbf{W}}_{N,g}}$ was the eigenbases of ${\frac{1}{M}}\sum\limits_{k=1}^{M}{{{{\mathbf{\bar{F}}}}_{N,k}}}\mathbf{\bar{F}}_{N,k}^{\operatorname{T}}$. Use mini-batch gradient descent to minimize the loss $\sum\nolimits_{{{I}_{k}}\in {{\beta }_{t}}}{\operatorname{tr}({{\mathbf{W}}_{N,g}}{{{\mathbf{\bar{F}}}}_{N,k}}\mathbf{\bar{F}}_{N,k}^{\operatorname{T}}\mathbf{W}_{N,g}^{\operatorname{T}})}/\left| {{\beta }_{t}} \right|$, where ${{\beta }_{t}}$ was a batch of sampled images at the t$^{th}$ iteration of the gradient descent. The use of the negative trace as a loss function introduced instability into the gradient descent process. These loss functions lacked a lower bound, causing gradient descent algorithms to prioritize minimizing losses while neglecting the constraint ${{\mathbf{W}}_{N,g}}\mathbf{W}_{N,g}^{\operatorname{T}}=\mathbf{I}$. This was unsolvable in the current situation, and we used Eq. \ref{eq4} rewriting as Eq. \ref{eq5} to solve and optimize the problem of ${{\mathbf{W}}_{N,g}}$. Among them, Eq. \ref{eq5} was approximately solvable in the case of small batch gradient descent, where $\mathbf{W}_{N,g}^{\operatorname{T}}{{\mathbf{W}}_{N,g}}{{\mathbf{\bar{F}}}_{N,k}}$ was the characteristic of ${{\mathbf{\bar{F}}}_{N,k}}$ according to the ${{\mathbf{W}}_{N,g}}{{\mathbf{\bar{F}}}_{N,k}}$ map.

%Eq5
\begin{equation}\label{eq5}
	\underset{{{\mathbf{W}}_{N,g}}\mathbf{W}_{N,g}^{\operatorname{T}}=\mathbf{I}}{\mathop{\max }}\,\frac{1}{M}\sum\limits_{k=1}^{M}{||\mathbf{W}_{N,g}^{\operatorname{T}}{{\mathbf{W}}_{N,g}}{{{\mathbf{\bar{F}}}}_{N,k}}-\mathbf{\bar{F}}_{N,k}^{{}}}||_{2}^{2}
\end{equation}

In summary, to obtain ${{\mathbf{W}}_{N,g}}(N=1,2,3,...,16)$, we employed a small batch gradient descent approach to simultaneously compute ${{\mathbf{W}}_{N,g}}$ in Eq. \ref{eq5}. During the $t^{th}$ iteration of the gradient descent, we sampled a batch denoted as ${{\beta }_{t}}$ for the following minimization problem, as presented in Eq. \ref{eq6}. We computed the gradient of the objective function to update ${{\mathbf{W}}_{N,g}}$. The batch size employed in this process was 8, and we trained ${{\mathbf{W}}_{N,g}}$ for 200 epochs.

Using global eigenbases ${{\mathbf{W}}_{N,g}}(N=1,2,3,...,16)$, we were able to transfer information from the $reluN$ layer of the source model $ENC$ to the $reluN{_{e}}$ layer of the target model $enc$, as illustrated in Figure \ref{fig2}(a).

%Eq6
\begin{equation}\label{eq6}
	\underset{\underset{N\in \{1,2,3,4\}}{\mathop{{{\mathbf{W}}_{N,g}}\mathbf{W}_{N,g}^{\operatorname{T}}=\mathbf{I}}}\,}{\mathop{\min }}\,\frac{1}{\left| {{\beta }_{t}} \right|}\sum\limits_{N=1}^{4}{\sum\limits_{{{I}_{k}}\in {{\beta }_{t}}}^{{}}{||\mathbf{W}_{N,g}^{\operatorname{T}}{{\mathbf{W}}_{N,g}}{{{\mathbf{\bar{F}}}}_{N,k}}-\mathbf{\bar{F}}_{N,k}^{{}}}||_{2}^{2}}
\end{equation}

\textbf{Blockwise PCA-based KD.} To facilitate feature transformation within the distillation model, it was necessary to incorporate a paired decoder denoted as $dec$. This decoder worked in tandem with the encoder $enc$ to extract input information effectively. The distillation method employed in this context was Principal Components Analysis (PCA). 

The encoder $enc$ was divided into a series of blocks $\left\{ enc{_1},enc{_2},enc{_3},...,enc{_{16}} \right\}$, where the output $en{{c}_{N}}$ was the $reluN{_{e}}$ layer, and the decoder $dec$ was divided into a group of blocks $\left\{ dec{_{16}},...,dec{_3},dec{_2},dec{_1} \right\}$, where the output $dec{_N}$ was the $relu(N-1){_d}$ layer with the reproduction characteristics $relu(N-1){_{e}}$. That was, the decoder took the $relu16{_{e}}$ features from $dec$ as input to progressively reproduce the $relu16{_{e}},...,relu3{_{e}},relu2{_{e}},relu1{_{e}}$ features and images. To implement our $enc-dec$ model, we trained each pair of $enc{_{N}}$ and $dec{_{N}}$ with other pairs minimizing the distillation loss $\mathcal L_{enc}^{N}$ of the encoder and the implementation loss $\mathcal L_{dec}^{N}$ of the decoder. These four pairs were trained in order from $N=1$ to $N=16$, as shown in Figure \ref{fig2}(b), with $enc{_{2}}$ and $dec{_{2}}$ as examples.

In the distillation of the encoder, given the image ${{I}_{k}}$, we wanted to train the encoder block $enc{_{N}}$ to make its decentralized output $\mathbf{\bar{F}}_{N,k}^{e}$ close to the feature ${{\mathbf{W}}_{N,g}}\mathbf{\bar{F}}_{N,k}$. It was derived from the $\mathbf{\bar{F}}_{N,k}$ map of the global eigenbases ${{\mathbf{W}}_{N,g}}$ from the decentralized output of $ENC{_{N}}$, with reconstruction loss as shown in Eq. \ref{eq7}.

%Eq7
\begin{equation}\label{eq7}
	\mathcal{L}_{enc}^{N}({{I}_{k}})=||\mathbf{W}_{N,g}^{\operatorname{T}}\mathbf{\bar{F}}_{N,k}^{e}-\mathbf{\bar{F}}_{N,k}||_{2}^{2}
\end{equation}

In the decoder implementation, given an image ${{I}_{k}}$, we wanted to approximate the output $\mathbf{F}_{N-1,k}^{d}$ of the $dec{_{N}}$ to the input image ${{I}_{k}}$ in order to reproduce the input $\mathbf{F}_{N-1,k}^{e}$ of the $enc{_{N}}$ and the reconstructed image from the $dec{_{1}}$. In general, we minimized the ${{I}_{k}}_{_{rec}}$ formed by three regular terms, as shown in Eq. \ref{eq8}.

%Eq8
\begin{equation}\label{eq8}
	\mathcal{L}_{dec}^{N}({{I}_{k}})=||\mathbf{F}_{N-1,k}^{d}-\mathbf{F}_{N-1,k}^{e}||_{2}^{2}+||{{I}_{{{k}_{rec}}}}-{{I}_{k}}||_{2}^{2}+||\mathbf{F}_{N,{{k}_{rec}}}^{{}}-\mathbf{F}_{N,k}^{{}}||_{2}^{2}
\end{equation}

The third objective was to enable image reconstruction based on perceptual loss. It was noteworthy that when $N=1$, the first term for feature reconstruction was absent. In summary, we had tackled the following optimization challenges during the training of $enc{{N}}$ and $dec{{N}}$:

%Eq9
\begin{equation}\label{eq9}
	\underset{enc{_{N}},dec{_{N}}}{\mathop{\min }}\,\mathcal{L}_{enc}^{N}({{I}_{k}})+\mathcal{L}_{dec}^{N}({{I}_{k}})
\end{equation}

\textbf{Reducing channel lengths.} In accordance with the empirical principles of PCA dimensionality reduction, it was imperative to preserve the most vital information encapsulated in the channel length $L_{N}^{e}$ of the target model $enc$. Specifically, the target layer $reluN{_{e}}$ of $L_{N}^{e}$ should retain variance information exceeding 85\% of that found in the source layer $reluN$.

For each image ${{I}_{k}}$ within our dataset, we calculated the covariance of its features (called $\mathbf{F}_{N,k}$) extracted in the $reluN$ layer. Let $\sigma _{N,k}^{j}$ be the ${{j}^{th}}$ largest eigenvalue of the covariance associated with the ${{j}^{th}}$ principal eigenvector $\mathbf{e}_{N,k}^{j}$. The ${{j}^{th}}$ explanatory variance, $EV=\sigma _{N,k}^{j}/\sum\nolimits_{\alpha =1}^{{{L}_{N}}}{\sigma _{N,k}^{\alpha }}$, reflected the portion of the feature variance captured by $\mathbf{e}_{N,k}^{j}$, while the cumulative explanatory variance, $CEV=\sum\nolimits_{\beta =1}^{L_{N}^{'}}{\sigma _{N,k}^{\beta }}/\sum\nolimits_{\alpha =1}^{{{L}_{N}}}{\sigma _{N,k}^{\alpha }}$, reflected the feature variance captured by the top $L_{N}^{'}$ feature vector. We used the mean cumulative explanatory variance (mCEV) to determine the value of $L_{N}^{e}$. The mCEV, represented as the mean of the CEV values across all images, was expressed in Eq. \ref{eq10}.

%Eq10
\begin{equation}\label{eq10}
	\operatorname{mCEV}(L_{N}^{'})=\frac{1}{M}\sum\limits_{k=1}^{M}{\frac{\sum\nolimits_{\beta =1}^{L_{N}^{'}}{\sigma _{N,k}^{\beta }}}{\sum\nolimits_{\alpha =1}^{L_{N}^{{}}}{\sigma _{N,k}^{\alpha }}}}=\sum\limits_{\beta =1}^{L_{N}^{'}}{\operatorname{mEV}(\beta )}
\end{equation}

\begin{equation}\label{eq10-2}
    \operatorname{mEV}(\beta )=\frac{1}{M}\sum\nolimits_{k=1}^{M}{(\sigma _{N,k}^{\beta }/\sum\nolimits_{\alpha =1}^{{{L}_{N}}}{\sigma _{N,k}^{\alpha }})}
\end{equation}

Where $M$ was the number of images in the dataset, and $\operatorname{mEV}(\beta)$ was the mean $\beta ^{th}$ EV. We were looking for $L_{N}^{e}$ that meets $\operatorname{mCEV}(L_{N}^{e})\approx$85\%.

% %Figure 2
% % \begin{figure*}[!t]
% \begin{figure*}
%         \renewcommand{\thefigure}{A2}
% 	\centering
% 	\includegraphics[width=6.5in]{Fig2}
% 	\caption{PCA-based knowledge distillation structure for VGG. Notes: PCA-based knowledge distillation consists of two steps: (a) global eigenbasis (${{\mathbf{W}}_{\mathbf{N}}},\mathbf{N}=1,2,3,...,16$) derivation, (b) blockwise PCA knowledge distillation, (c) the basic autoencoder framework of VGG-19, and (d) the $enc-dec$ autoencoder framework from our method. $relu16$ = $relu16'$, and $relu16{_e}$ = $relu16{_d}$.}
% 	\label{fig2}
% \end{figure*}

\section{Implicit Diffusion model}
\label{section_idm}

\textbf{Diffusion process.} The diffusion process referred to the process of gradually adding Gaussian noise to the data until the data became random noise. For the original data ${{\mathbf{x}}_{0}}\pi \sim q({{\mathbf{x}}_{0}})$, each step of the diffusion process with a total of $T$ steps was to add Gaussian noise to the data ${{\mathbf{x}}_{t-1}}$ obtained in the previous step as follows: 

%Eq13
\begin{equation}\label{eq13}
	q({{\mathbf{x}}_{t}}|{{\mathbf{x}}_{t-1}})=\mathcal{N}({{\mathbf{x}}_{t}};\sqrt{1-{{\beta }_{t}}}{{\mathbf{x}}_{t-1}},{{\beta }_{t}}\mathbf{I})
\end{equation}

Where $\{{{\beta }_{t}}\}_{t=1}^{T}$ was the variance used for each step, which was between 0 and 1. For the diffusion model, we often called the variance of different steps the variance schedule or noise schedule. Usually, the latter step would adopt a larger variance, that was ${{\beta }_{1}}<{{\beta }_{2}}<...<{{\beta }_{T}}$. Under a designed variance schedule, if the number of diffusion steps was large enough, the final result ${{\mathbf{x}}_{T}}$ would completely lose the original data and become random noise. Each step of the diffusion process generated noisy data ${{\mathbf{x}}_{t}}$. The whole diffusion process was also a Markov chain:

%Eq14
\begin{equation}\label{eq14}
	q({{\mathbf{x}}_{1:T}}|{{\mathbf{x}}_{0}})=\prod\limits_{t=1}^{T}{q({{\mathbf{x}}_{t}}|{{\mathbf{x}}_{t-1}})}
\end{equation}

\textbf{Reverse process.} 
The diffusion process entailed the introduction of noise to data, whereas the reverse process served the purpose of noise removal. In the context of the reverse process, the knowledge of the true distribution $q({{\mathbf{x}}_{t-1}}|{{\mathbf{x}}_{t}})$ at each step was crucial. Starting from an initial random noise ${{\mathbf{x}}_{T}}\sim \mathcal{N}(0,\mathbf{I})$, the gradual denoising procedure culminated in the generation of a genuine sample. Consequently, it was evident that the reverse process also corresponded to the data generation process.

%Table 1
% \begin{table*}[!t]
% \begin{table*}[!b]
\begin{table}[!ht]
        \renewcommand{\thetable}{A3}
	\caption{Size and inference time of different models. Notes: DDPM = Denoising Diffusion Probabilistic Models and DDIM = Denoising Diffusion Implicit Models. The experiment was implemented in NVIDIA TESLA V100 32GB GPU. Size and inference time of different models. The latent dims are 64, the number of inference steps is 20, and the image size is 256. SD = Stable diffusion. Units: Parameter size (MB), Inference time (second/image) and KD ratio (\%). \label{tab:table1}}
	\centering
        \resizebox{1.0\textwidth}{!}{
	\begin{tabular}{ccccc}
		\toprule
		\textbf{Model} & \textbf{Trainable params} & \textbf{Params size} & \textbf{Inference time} & \textbf{KD ratio} \\ \midrule
		VGG-11 & 9,220,993 & 35.18  & 1.80$\times10^4$ & - \\ 
		VGG-16 & 14,715,201 & 56.13  & 2.54$\times10^4$ & - \\ 
		VGG-19 & 20,024,897 & 76.39  & 2.89$\times10^4$ & - \\ 
		UNet & 31,037,698 & 118.48  & 8.43$\times10^2$ & - \\ 
		KD-VGG-11 & 49,897 & 0.19  & 1.75$\times10^4$ & 99.46  \\ 
		KD-VGG-16 & 164,425 & 0.63  & 2.19$\times10^4$ & 98.88  \\ 
		KD-VGG-19 & 260,100 & 0.99  & 2.60$\times10^4$ & 98.70  \\ 
		KD-UNet & 14,673,253 & 55.97  & 5.22$\times10^2$ & 52.72  \\ 
		VAE & 52,519,176 & 12.50  & 1.21$\times10^3$ & - \\ 
		GAN & 67,876,224 & 258.95  & 3.93$\times10^2$ & - \\ 
		SD-V1.4 & 1,370,661,847 & 5228.66  & 1.07  & - \\ 
		SD-V1.5 & 1,370,661,847 & 5228.66  & 1.10  & - \\ 
		SD-V2.1 & 1,290,388,111 & 4922.44  & 1.01  & - \\ 
		DDPM & 113,712,824 & 433.78  & 0.96  & - \\ 
		IDM & 116,669,808 & 445.06  & 0.97  & - \\ 
		Ours-VGG-11 & 97,376,010 & 371.46  & 0.79  & 16.54  \\ 
		Ours-VGG-16 & 97,491,353 & 371.90  & 0.79  & 16.44  \\ 
		Ours-VGG-19 & 97,585,725  & 372.26  & 0.80  & 16.36  \\ \bottomrule
	\end{tabular}
    }
\end{table}

%Figure 5
\begin{figure*}[!ht]
        \renewcommand{\thefigure}{A1}
	\centering
	\includegraphics[width=1.0\textwidth]{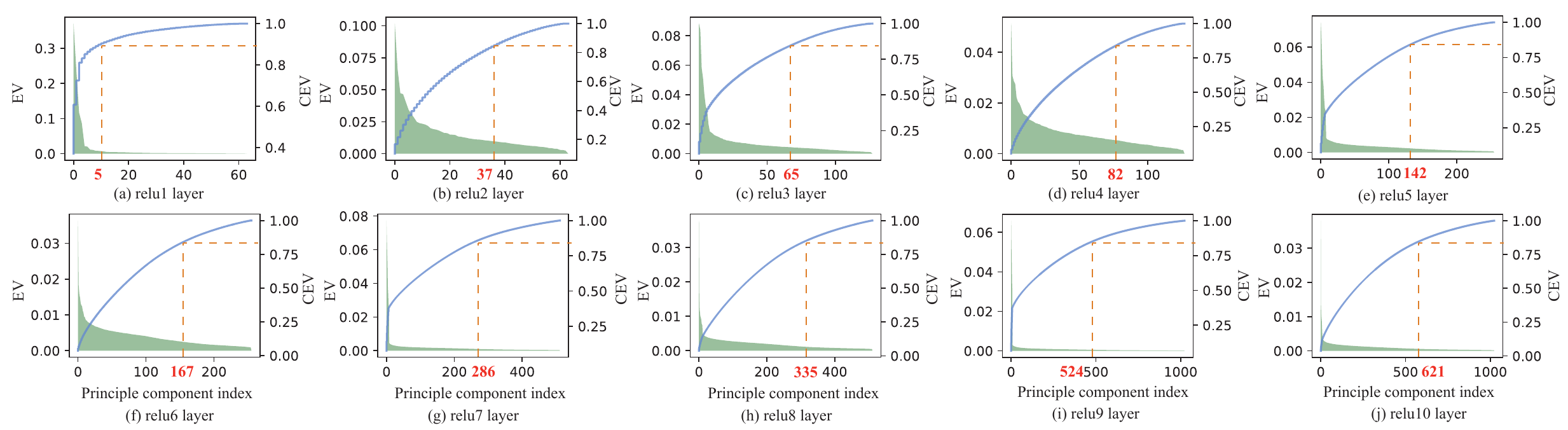}
	\caption{Features of UNet. Notes: Mean explained variance (green area) and mean cumulative explained variance (blue curve) of the $reluN$ features.} \label{fig5}
\end{figure*}

\end{document}